\documentclass[10pt]{article} 
\usepackage[accepted]{tmlr}
\usepackage{natbib} 
\usepackage{mathtools} 
\usepackage{booktabs} 
\usepackage{tikz} 

\usepackage{hyperref}
\usepackage{url}
\usepackage[utf8]{inputenc} 
\usepackage[T1]{fontenc}    
\usepackage{hyperref}       
\usepackage{url}            
\usepackage{booktabs}       
\usepackage{nicefrac}       
\usepackage{microtype}      
\usepackage{xcolor}         
\usepackage{microtype}
\usepackage{graphicx}
\usepackage{overpic}
\usepackage{adjustbox}
\usepackage{subfigure}
\usepackage{multirow}
\usepackage{booktabs} 
\usepackage{multirow}
\usepackage{enumitem}
\usepackage{amsmath,algorithm}
\usepackage{anyfontsize}
\usepackage{soul}
\usepackage{float}
\usepackage{lipsum} 

\usepackage{algorithm}
\usepackage{algorithmic}

\usepackage{amsfonts}       
\usepackage{amsmath}
\usepackage{amssymb}
\usepackage{mathtools}
\usepackage{amsthm}

\usepackage{amsmath,amsfonts,bm}

\def\eqref#1{equation~\ref{#1}}

\def\1{\bm{1}}

\DeclareMathAlphabet{\mathsfit}{\encodingdefault}{\sfdefault}{m}{sl}
\SetMathAlphabet{\mathsfit}{bold}{\encodingdefault}{\sfdefault}{bx}{n}

\usepackage{hyperref}
\usepackage{url}

\let\classAND\AND
\let\AND\relax
\usepackage{algorithmic}

\let\AND\classAND
\AtBeginEnvironment{algorithmic}{\let\AND\algoAND}

\hypersetup{
    colorlinks=true,       
    linkcolor=blue,        
    citecolor=magenta,       
    filecolor=magenta,     
    urlcolor=cyan,         
    unicode=true,           
    breaklinks=true,        
    plainpages=false       
}

\title{Data Augmentation Policy Search for Long-Term Forecasting}

\author{\name Liran Nochumsohn \email lirannoc@post.bgu.ac.il \\
      \addr Department of Computer Science\\
        Ben-Gurion University of the Negev
      \AND
      \name Omri Azencot \email azencot@bgu.ac.il \\
      \addr Department of Computer Science\\
        Ben-Gurion University of the Negev}

\def\month{02}  
\def\year{2025} 
\def\openreview{\url{https://openreview.net/forum?id=Wnd0XY0twh&noteId=ObOhmTtATp}} 

\begin{document}

\maketitle

\begin{abstract}
Data augmentation serves as a popular regularization technique to combat overfitting challenges in neural networks. While automatic augmentation has demonstrated success in image classification tasks, its application to time-series problems, particularly in long-term forecasting, has received comparatively less attention. To address this gap, we introduce a time-series automatic augmentation approach named TSAA, which is both efficient and easy to implement. The solution involves tackling the associated bilevel optimization problem through a two-step process: initially training a non-augmented model for a limited number of epochs, followed by an iterative split procedure. During this iterative process, we alternate between identifying a robust augmentation policy through Bayesian optimization and refining the model while discarding suboptimal runs. Extensive evaluations on challenging univariate and multivariate forecasting benchmark problems demonstrate that TSAA consistently outperforms several robust baselines, suggesting its potential integration into prediction pipelines. Code is available at this repository: \href{https://github.com/azencot-group/TSAA}{https://github.com/azencot-group/TSAA}.
\end{abstract}

\section{Introduction}

Modern machine learning tools require large volumes of data to effectively solve challenging tasks. However, high-quality labeled data is difficult to obtain as manual labeling is costly and it may require human expertise~\citep{shorten2019survey}. Small datasets may lead to overfitting in overparameterized models, a phenomenon in which the model struggles with examples it has not seen before~\citep{allen2019learning}. One of the effective methods to alleviate poor generalization issues is via \emph{data augmentation} (DA). Data augmentation aims to generate artificial new examples whose statistical features match the true distribution of the data~\citep{simard1998transformation}. In practice, DA has been shown to achieve state-of-the-art (SOTA) results in e.g., vision~\citep{krizhevsky2012imagenet} and natural language~\citep{wei2019eda} tasks.

Unfortunately, DA is not free from challenges. For instance, \cite{tian2020makes} showed that the effectivity of augmented samples depends on the downstream task. To this end, recent approaches explored automatic augmentation tools, where a good DA policy is searched for~\citep{lemley2017smart, cubuk2019autoaugment}. While automatic frameworks achieved impressive results on image classification tasks~\citep{zheng2022deep} and other data modalities, problems with time-series data received significantly less attention~\cite{kaufman2024first, nochumsohn2024beyond}. Toward bridging this gap, we propose in this work a new automatic data augmentation method, designed for \emph{time-series forecasting} problems.

Time-series forecasting is a long-standing task in numerous scientific and engineering fields~\citep{chatfield2000time}. While deep learning techniques achieved groundbreaking results on vision and NLP problems already a decade ago, time-series forecasting (TSF) was considered by many to be too challenging for deep models, up until recently~\citep{oreshkin2019n}. While recent linear approaches showed interesting forecast results~\citep{zeng2022transformers,nochumsohn2025multitasklearningapproachlinear}, existing SOTA approaches for TSF are based on deep learning architectures that are structurally similar to vision models. In particular, current TSF deep models are overparameterized~\cite{kaufman2024geometric}, and thus they may benefit from similar regularization techniques which were found effective for vision models, such as (automatic) data augmentation. Ultimately, our work is motivated by the limited availability of DA tools for time-series tasks~\citep{wen2020time}.


The main contributions of our work can be summarized as follows:
1) We develop a novel automatic data augmentation approach for long-term time-series forecasting tasks. Our approach is based on a carefully designed dictionary of time-series transformations, Bayesian optimization for policy search, and pruning tools that enforce early stopping of ineffective networks. While these components appear in existing work, their combination and adaptation to time-series forecasting was not done before, to the best of our knowledge.
2) We analyze the optimal policies our approach finds. Our analysis sheds light into the most effective transformations, and it may inspire others in designing effective data augmentation techniques for time-series data.
3) Our approach augments existing time-series forecasting baselines, and we extensively evaluated it on long-term forecasting univariate and multivariate TSF benchmarks with respect to several strong baseline architectures. We find that our framework enhances performance in most long-term forecast settings and across most datasets and baseline architectures.

\section{Related Work}

\paragraph{Time-series forecasting.} Recently, several neural network approaches for TSF have been proposed. Based on recurrent neural networks, DeepAR~\citep{salinas2020deepar} produced probabilistic forecasts with uncertainty quantification. The N-BEATS~\citep{oreshkin2019n} model employs fully connected layers with skip connections, and subsequent work \citep{challu2022n} improved long-term forecasting via pooling and interpolation. Another line of works based on the transformer architecture~\citep{vaswani2017attention} used a sparse encoder and a generative decoder in the Informer \citep{zhou2021informer}, trend-seasonality decomposition in the Autoformer~\citep{wu2021autoformer}, and Fourier and Wavelet transformations in the FEDformer~\citep{zhou2022fedformer}. Recently, Pyraformer~\citep{liu2021pyraformer} significantly reduced the complexity bottleneck of the attention mechanism, PatchTST~\citep{nie2022time} exchanges the point-wise attention input with a tokenized sub-series representation. Finally, \cite{zeng2022transformers} propose a single-layer MLP with a larger input lookback.

\begin{figure}[t]
  \centering
  \begin{overpic}[width=1\linewidth]{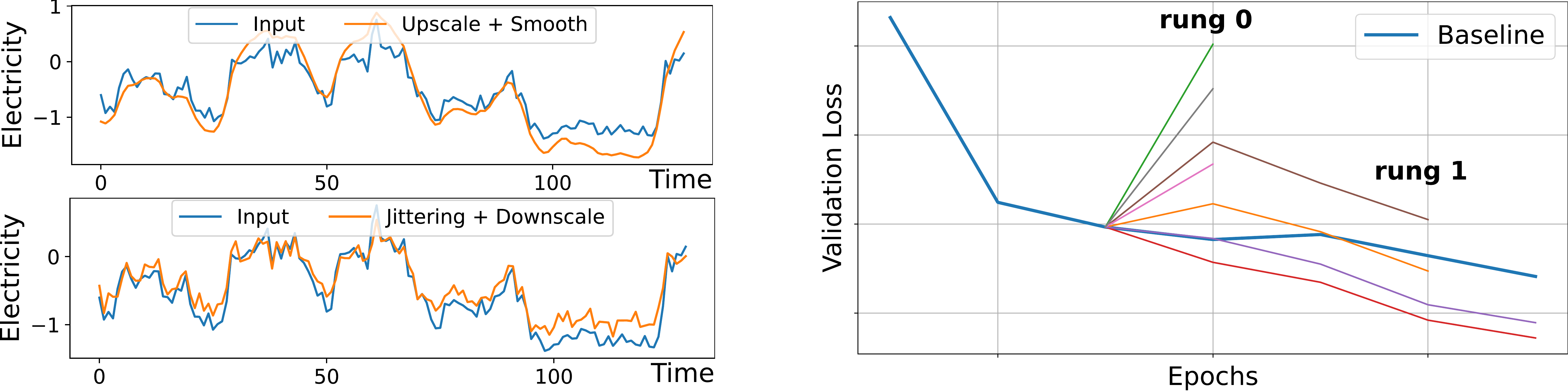} 
        \put(1, 0){A} \put(51.5, 0){B}
  \end{overpic}
  \caption{A) Two examples of sub-policies applied on Electricity data. B) The above plot demonstrates the behavior of ASHA with respect to the baseline model (blue). Some of the poorly performing runs are discontinued at the end of rungs, whereas the other runs train to completion.}  
  \label{fig:ts_asha_policy}
\end{figure}

\paragraph{Data augmentation.} DA techniques have appeared since the early rise of modern deep learning to promote labeled image invariance to certain transformations~\citep{krizhevsky2012imagenet}. Typical image augmentations include rotation, scaling, crop, and color manipulations. Recent methods focused on modality-agnostic methods which blend linearly the inputs and labels~\citep{zhang2017mixup} or utilize manifold learning approaches~\cite{kaufman2023data, kaufman2024geometric, kaufman2024first}. Other works produce augmented views in the feature space~\citep{devries2017dataset, verma2019manifold}. In contrast to image and text data, augmenting arbitrary time-series (TS) data have received less attention in the literature~\citep{wen2020time, iwana2021empirical}. In the review~\citep{wen2020time}, the authors consider three different tasks: TS classification, TS anomaly detection, and TS forecasting. Their analysis is based on common time-series augmentation approaches such as scaling, adding noise~\citep{um2017data}, window cropping or slicing, and stretching of time intervals~\citep{le2016data}, dynamic time warping~\citep{ismail2019deep}, perturbations of the frequency domain~\citep{gao2020robusttad, chen2023fraug}, and utilizing surrogate data~\citep{lee2019surrogate}. In~\citep{smyl2016data}, the authors discuss additional TS augmentation approaches including generating new TS using the residuals of a statistical TS~\citep{bergmeir2016bagging}. Another technique would be to sub-sample the parameters, residuals, and forecasting from MCMC Bayesian models. The survey~\citep{iwana2021empirical} further details a large list of TS augmentations such as jittering, rotation, time warping, time masking, interpolation and others in the context of time-series classification. The authors in~\citep{wen2020time} propose the selection and combination of augmentations using automatic approaches as a promising avenue for future research, which is the focus of the current work. Finally, we also mention the large body of work on generative modeling of time series~\citep{yoon2019time, naiman2024generative, naiman2024utilizing} which is related to data augmentation. We show in Fig.~\ref{fig:ts_asha_policy}A two examples of DA policies.

\paragraph{Automatic DA.} To avoid hand-tailored DA, recent efforts aimed for automatic tools, motivated by similar advances in neural architecture search (NAS) approaches~\citep{zoph2016neural}. AutoAugment~\citep{cubuk2019autoaugment} used a recurrent controller along with reinforcement learning for the search process, yielding a highly effective but computationally intensive framework. Following works such as Fast AutoAugment employed Bayesian optimization and density matching~\citep{lim2019fast}. RandAugment \citep{cubuk2020randaugment} reduces the search space significantly by introducing stochasticity. \cite{tian2020improving} suggested partial training using augmentation-wise weight sharing (AWS). Further, recent approaches utilize gradients for the search problem, including the differentiable automatic DA (DADA)~\citep{li2020dada} and Deep AutoAugment~\citep{zheng2022deep}. \cite{cheung2020modals} developed automatic DA that does not depend on the data-modality as it exploits latent transformations. In~\citep{fons2021adaptive}, the authors propose adaptive-weighting strategies which favor a subset of time-series DA for classification, based on their effect on the training loss.

\section{Background}
\label{sec:background}

Below, we briefly describe background information on Bayesian optimization and pruning approaches which we use to find the best augmentation policy and improve model training efficiency, respectively.

\paragraph{Tree-structured Parzen Estimators and the Expected Improvement.} Bayesian optimization relates to a family of techniques where an objective function $f(x): \mathbb{R}^d \rightarrow \mathbb{R}^+$ is minimized, i.e.,
\begin{equation} \label{eq:bayes_opt}
    \min_x f(x) \ .
\end{equation}
In the typical setting, $f$ is costly to evaluate, its gradients are not available, and $d \leq 20$. For instance, finding the hyperparameters ($x$) of a neural network ($f$) is a common use case for Bayesian optimization~\citep{bergstra2013making}. Unlike grid/random search, Bayesian optimization methods utilize past evaluations of $f$ to maintain a surrogate model  $p(y|x)$  for the objective function $y=f(x)$. Thus, Bayesian optimization solves Eq.~\ref{eq:bayes_opt} while limiting the costly evaluations of $f$ to a minimum.

A practical realization of Bayesian optimization is given by Sequential Model-Based Optimization (SMBO)~\citep{hutter2011sequential}. SMBO iterates between model fitting with the existing parameters (exploitation) to parameter selection using the current model (exploration). SMBO constructs a surrogate model $p(y|x)$, finds a set of parameters $x$ that performs best on the $p(y|x)$ using an acquisition function, applies the objective function $f$ on $x$ to obtain the score $y$, updates the surrogate model, and repeats the last three steps until convergence. Most SMBO techniques differ in their choice of the surrogate model and acquisition function. We will focus on Tree-structured Parzen Estimator (TPE) for the surrogate model, combined with Expected Improvement for the acquisition function. The main idea behind TPE is to model the surrogate via two distributions, $l(x)$ and $g(x)$, corresponding to model evaluations that yield positive, and negative improvement. Formally, 
\begin{equation}
    p(x|y) = 
    \begin{cases}
    l(x) \quad y < y^* \\
    g(x) \quad y \ge y^*
    \end{cases} \ ,
\end{equation}
where $y^*$ is a threshold score, and the surrogate model is obtained via Bayes rule. It can be shown that maximizing $l(x) / g(x)$ leads to an optimal Expected Improvement (EI)~\citep{bergstra2011algorithms}.

\paragraph{Asynchronous Successive Halving.} While Bayesian optimization uses a minimal number of evaluations of $f$, the overall minimization is computationally demanding due to the high cost of $f$, e.g., if $f$ is a neural network that needs to be trained. To alleviate some of these costs, Asynchronous Successive Halving (ASHA)~\citep{jamieson2016non, li2020system} enforces early stopping of poorly performing parameters $x$, whereas parameters with low $l(x)$, are trained to the fullest. In a fixed budget system, given a maximum resource $R$, minimum resource $r$, and a reduction factor $\eta$, ASHA works as follows. One creates model checkpoints during the training process at epochs $\eta^j$ where $j=1,\dots , \lfloor \log_\eta R/r \rfloor $. Each checkpoint is referred to as a \emph{rung}, and at the end of each rung, one keeps only the best $\frac{1}{\eta}$ runs. To avoid waiting for all runs to reach the next rung, ASHA performs asynchronous evaluations to promote or halt runs on the go. We illustrate in Fig.~\ref{fig:ts_asha_policy}B an example of a baseline model with multiple different runs, administered by ASHA.

\section{Time-Series AutoAugment (TSAA)}
\label{sec:method}

\paragraph{Automatic augmentation via bi-level optimization.} The task of finding data augmentations automatically during the training of a deep neural network model can be formulated as a bi-level optimization problem, see e.g., (Li et al., 2020b). Namely, 
    \setcounter{equation}{2}
    \begin{align}
    \min_\theta \quad & \mathcal{L}_\text{val} (\omega, \theta)  \label{eq:bilevel_top} \\
    \text{subject to} \quad & \min_\omega \mathbb{E}_{p_\theta} \left[ \mathcal{L}_\text{tr}(\omega, \theta) \right] \label{eq:bilevel_bot} \ .
\end{align}
At the top level, Eq.~\ref{eq:bilevel_top}, the optimization aims to find the optimal augmentation policy $\theta \sim p_\theta$, where $p_\theta$ is some distribution of augmentation policies, e.g., additive noise. Importantly, Eq.~\ref{eq:bilevel_top} minimizes the validation loss, $\mathcal{L}_\text{val}$, that is parameterized both by the augmentation policy and by the network weights $\omega$. The latter weights are obtained in the bottom level, Eq.~\ref{eq:bilevel_bot}, describing an optimization problem that is similar to standard training of a neural network as it minimizes the training loss, $\mathcal{L}_\text{tr}$. The main two differences in Eq.~\ref{eq:bilevel_bot} from standard training is the dependence on $\theta$ and the expectation over all possible augmentation policy distributions $p_\theta$, leading overall to mutually-dependent optimization problems, or, a bi-level optimization. Unfortunately, the above problem is difficult to solve in practice, and therefore, we relax it as detailed next.

\begin{figure*}[t]
  \centering
  \begin{overpic}[width=1\linewidth]{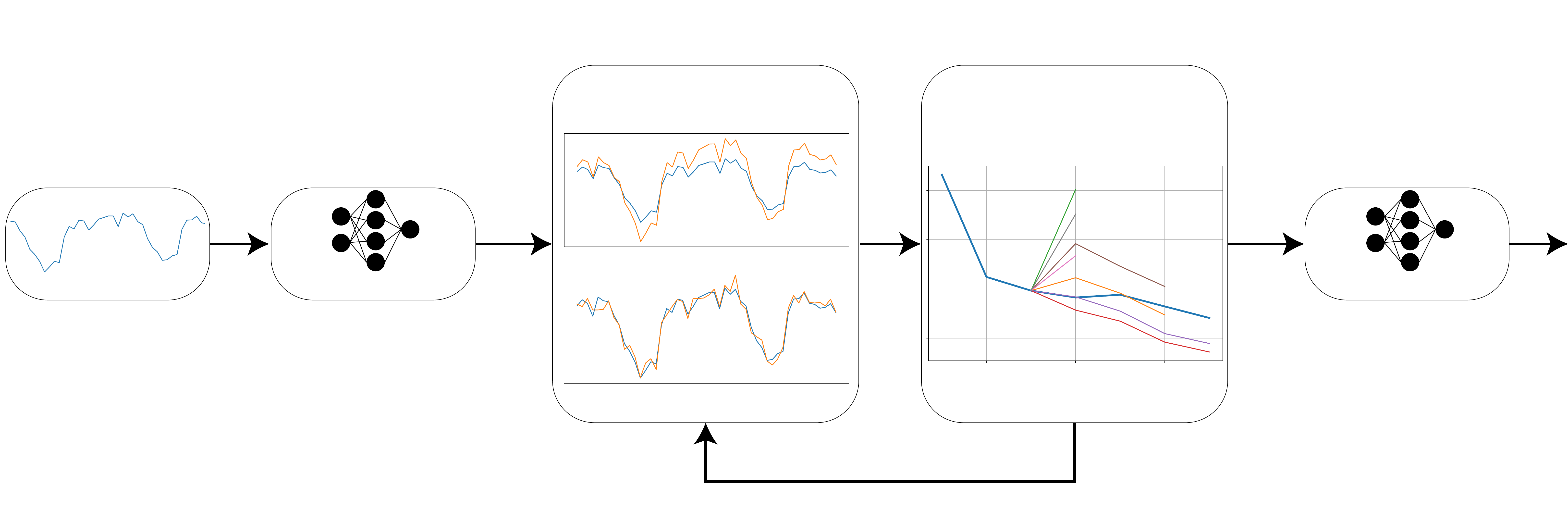} 
        \put(4, 23){\fontsize{12}{14.4}\selectfont input}
        \put(20, 23){\fontsize{12}{14.4}\selectfont Step 1} \put(30.5, 20){\small $\omega_\text{share}$}
        \put(53.5, 31){\fontsize{12}{14.4}\selectfont Step 2} \put(56, 20){$\theta$} \put(56, 4){$\omega$}
        \put(36, 25){\scriptsize Seasonality Upscale} \put(36, 16.5){\scriptsize Jitter}
        \put(42, 27){TPE} \put(59.5, 27){fine-tune + ASHA}
        \put(79, 20){$p_{\theta^*}$} \put(79, 14){$\theta^*$} \put(97, 20){$\omega^*$}
        \put(18.5, 14.5){\small partial train} \put(86, 14.5){\small fine-tune}
        \put(40, 7){\small find policy} \put(63, 7){\small find weights}
  \end{overpic}
  \caption{Our time-series automatic augmentation (TSAA) approach is based on a partial train of the model (Step $1$), followed by an iterative process (Step $2$) where we alternate between improving the augmentation policy  $\theta$ to training the model weights $\omega$. We find $\omega^*$ by fine-tuning over $p_{\theta^*}$.}  
  \label{fig:method_arch}
\end{figure*}

\paragraph{TSAA overview.} Our approach, which we call time-series automatic augmentation (TSAA), consists of two main steps, as illustrated in Fig.~\ref{fig:method_arch} and summarized in Alg.~\ref{alg:tsaa}. In the first step, we partially train the model for a few epochs and construct a set of shared weights. The second step iterates between solving Eq.~\ref{eq:bilevel_top} in search of an augmentation policy using TPE and EI to solving Eq.~\ref{eq:bilevel_bot} with fine-tuning and ASHA for an optimal model. A complexity analysis is given in App.~\ref{app:complexity}.

\paragraph{Step 1: compute shared weights.} Solving Eq.~\ref{eq:bilevel_bot} iteratively requires repeated trainings of the deep model, which is computationally prohibitive. To reduce these costs, we propose to partially train the baseline model and generate a shared set of weights $\omega_\text{share}$. Doing so, Step 2 is reduced to an iterative process of fine-tuning models for a small number of epochs, where $\omega_\text{share}$ are shared across all augmentations policies. Beyond efficiency aspects, applying DA in the later stages of training is assumed to be more influential~\citep{tian2020improving}. In practice, we partially train for $\lfloor {\beta K} \rfloor$ epochs, where $\beta=0.5$ is a hyperparameter and $K$ is the \emph{active} number of train epochs. In our tests, $K\leq 10$, and it may be strictly less due to an early stopping scheduler. $K$ is found by training the baseline model with no augmentation to completion and saving the weights after every epoch. Then, we define
\begin{equation*}
    \omega_\text{share} := \omega(\lfloor {\beta K} \rfloor) \ , \quad R := K - \lfloor {\beta K} \rfloor\ , 
\end{equation*}
where $R$ is the maximum resource parameter, and $r=1$ is the minimum resource, see Sec.~\ref{sec:background}.

\paragraph{Step 2: iterative split optimization.} Given $\omega_\text{share}$, it remains to solve Eqs.~\ref{eq:bilevel_top} and \ref{eq:bilevel_bot} to find the best augmentation policy $\theta^*$ and final weights $\omega^*$. In TSAA, we propose to split this problem to an iterative process, where we alternate between exploring augmentation policies $\theta$ via Eq.~\ref{eq:bilevel_top} to exploiting the current policy and produce model weights $\omega$ via Eq.~\ref{eq:bilevel_bot}. Namely, for a fixed set of weights $\omega$, the upper minimization finds the next policy $\theta$ to try by evaluating the validation set. Then, we fine-tune the model using a fixed $\theta$ with early stopping for a maximum of $R$ epochs to produce the next $\omega$. This procedure is repeated until a predefined number of trials $T_\text{max}$ is reached. The $k$ best-performing policies define $p_{\theta^*}$ from which $\theta^*$ is sampled, where we only allow policies that improve the baseline validation loss. Finally, we fine-tune the model again to obtain $\omega^*$. 

\paragraph{Solving Eq.~\ref{eq:bilevel_top}.} Existing work solved the upper problem using reinforcement learning~\citep{cubuk2019autoaugment, tian2020improving}, grid search~\citep{cubuk2020randaugment, fons2021adaptive}, and one-pass optimization~\citep{li2020dada, zheng2022deep}. Inspired by~\citep{lim2019fast}, we propose to use Tree-structured Parzen Estimator (TPE) with Expected Improvement (EI), see Sec.~\ref{sec:background}. In the context of TSAA, the parameters $x$ in Eq.~\ref{eq:bayes_opt} represent the policy $\theta$ and $f$ is $\mathcal{L}_\text{val}$. The Bayesian optimization is conducted over the policy search space and time-series augmentations we describe below.

\paragraph{Policy search space.} The augmentation policies $\theta$ we consider are drawn from a distribution $p_\theta$ over $k$ sub-policies $\Theta = \{\theta_1, \dots, \theta_k\}$. That is, the distribution $p_\theta$ aggregates several independent distributions $p(\theta_j)$, where $\theta_j \in \Theta$. Thus, $p_\theta$ allows to sample an augmentation $\theta$ from each of the considered $p(\theta_j)$, i.e.,
\begin{equation}
    \theta \sim p_\theta := \prod_{j=1}^k p(\theta_j) \ , \quad \theta_j \in \Theta \ , \Theta := \{\theta_1, \dots, \theta_k\} \ .
\end{equation}
Each sub-policy $\theta_j$ is composed of $n$ transformations $T_{j, i}$, applied sequentially on the output data $x_{i-1}$ of the previous transformation with $x_0$ being the input data and $m_{j, i}$ being the magnitude of the transformation. For instance, if the sub-policy $\theta_j$ consists of trend downscale in $T_{j, 1}$ and flip in $T_{j, 2}$ with magnitudes $m_{j, 1}= \frac{1}{2}, m_{j, 2}= \frac{1}{4}$, respectively, then, $\theta_j = T_{j, 2}[T_{j, 1}(x_0, \frac{1}{2}), \frac{1}{4}]$. For the general case of $n$ transformations, $\theta_j$ is defined via
\begin{equation}
    \theta_j = T_{j, n}(x_{n-1}, m_{j, n}) \circ \dots \circ T_{j, 1}(x_0, m_{j, 1}) \ .
\end{equation}

\paragraph{Time-series data augmentations.} While natural images are invariant to geometric transformations as translation and rotation, arbitrary time-series data need not be invariant to a certain type of transformations. Moreover, capturing the invariance in regression problems such as TSF may be more challenging than in classification tasks including images~\citep{kaufman2024first}. Finally, time-series data may include slow and fast phenomena such as bursts of electricity usage and seasonal peaks, for which some DA may be inapplicable. Thus, we propose to exploit DA that manipulate some features of the data and leave some features unchanged. For example, adjusting the trend while keeping the seasonality and noise components unaffected, or diversifying the time intervals in a way that the series mean and variance still stay the same. In particular, we suggest the following time-series transformations: identity, jittering, trend scaling, seasonality scaling, scaling, smoothing, noise scaling, flip, permutation, reverse, dynamic-time-stretching (DTS), window warping, and mixup. The magnitude of the augmentations can be controlled using a single parameter. The transformations are further elaborated in App.~\ref{app:ts_da} and Tab.~\ref{tab:ts_da} in the appendix.

\paragraph{Solving Eq.~\ref{eq:bilevel_bot}.} Finally, solving the bottom minimization may be achieved in a straightforward fashion via fine-tuning. However, as motivated in Sec.~\ref{sec:method}, doing so iteratively is costly. To prune runs, we augment our approach with Asynchronous Successive Halving (ASHA). Our choice to use ASHA over other techniques such as Bayesian Optimization HyperBand (BOHB)~\citep{falkner2018bohb} is motivated by the following reasons. First, BOHB has shown to be slightly inferior to ASHA~\citep{li2020system}. Second, In our setting $R \in \{1,2,..,5\}$ and  $\eta$ is set to be more aggresive. As a result, only two SHA brackets at most can be exploited in the HyperBand, thus limiting its effectiveness.

\label{pseudocode}
\vspace{-5mm}
\begin{center}
\begin{minipage}{0.75\textwidth}
\begin{algorithm}[H]
    \small
    \caption{Time-Series AutoAugment (TSAA)}
    \label{alg:tsaa}
    \begin{algorithmic}
       \STATE {\bfseries Inputs:} partial train factor $\beta$, resources $R, r$, max trials $T_\text{max}$, reduction factor $\eta$, and $k$ best DA sub-policies
       \STATE 
       \STATE $\Lambda \leftarrow \emptyset$  \hspace{\fill}\COMMENT {empty set into $\Lambda$}
       
       \STATE $\omega_\text{share}, R \leftarrow$ partial solve Eq. \ref{eq:bilevel_bot} with $\beta$

       STATE $\omega_0 \leftarrow \omega_\text{share}$ \hspace{\fill} \COMMENT{initialize the weights $\omega_0$ using the partial solution}
              
       \FOR{$i=1$ {\bfseries to} $T_\text{max}$} 
       
        \STATE $\theta_i \leftarrow$ solve Eq.~\ref{eq:bilevel_top} with $\text{TPE}(\Theta, \omega_{i-1})$
       
        \STATE $w_i \leftarrow$ fine-tune Eq.~\ref{eq:bilevel_bot} with $\omega_\text{share}$ and $\text{ASHA}(r,R,\eta)$ 
       
       \STATE $\Lambda \leftarrow \Lambda \cup  \bigl\{ [ \theta_i, \; \mathcal{L}_\text{val} (\omega_i, \theta_i) ] \bigl\}$ \hspace{\fill}\COMMENT {add found policy and loss to $\Lambda$}
       \ENDFOR
       \STATE $p_{\theta^*} \leftarrow k$ best sub-policies $\theta_i$ from $\Lambda$ 
       \STATE $\omega^* \leftarrow$ fine-tune $\mathcal{L}_\text{tr}(\theta^* \sim p_{\theta^*})$
       \STATE {\bfseries return} $p_{\theta^*}, \theta^*, \omega^*$ \hspace{\fill}\COMMENT {return the optimal DA distribution, policy, and network weights}
    \end{algorithmic}
\end{algorithm}
\end{minipage}
\end{center}

\section{Results}
\label{sec:results}

In what follows, we provide a comprehensive overview of our experimental setup, including models, datasets, and implementation details followed by evaluations of our approach. In the supplementary material, we offer additional information on hyperparameters (App.~\ref{app:hyperparam}), and extended results (App.~\ref{app:ex_results}).

\subsection{Models and datasets}
\label{app:model_data}

We extensively evaluate the performance of our Time-Series AutoAugment (TSAA) framework. To this end, we selected some of the most recent prominent time-series forecasting models. We consider the baseline architectures: \textbf{N-BEATS}~\citep{oreshkin2019n}, a deep neural architecture based on backward and forward residual links and a very deep stack of fully-connected layers. \textbf{Informer}~\citep{zhou2021informer} adapts the Transformer~\citep{vaswani2017attention} architecture to time-series forecasting tasks, with a new attention mechanism. \textbf{Autoformer}~\citep{wu2021autoformer} exchanges the self-attention module for an auto-correlation mechanism and introduces time-series decomposition as part of the model's encoding. Finally, \textbf{FEDformer}~\citep{zhou2022fedformer} enables capturing more important details in time-series through frequency domain mapping.

For each of the given baseline models, we apply TSAA on six commonly-used datasets in the literature of long-term time-series forecasting: (1) \textbf{ETTm2} \citep{zhou2021informer} contains electricity transformer oil temperature data alongside $6$ power load features. (2) \textbf{Electricity} \citep{zhou2021informer} is a collection of hourly electricity consumption data over the span of $2$ years. (3) \textbf{Exchange} \citep{lai2018modeling} consists of $17$ years of daily foreign exchange rate records representing different currency pairs. (4) \textbf{Traffic} \citep{zhou2021informer} is an hourly reported sensor data containing information about road occupancy rates. (5) \textbf{Weather} \citep{zhou2021informer} contains $21$ different meteorological measurements, recorded every $10$ minutes for an entire year. (6) \textbf{ILI} \citep{wu2021autoformer} includes weekly recordings of influenza-like illness patients.

We summarize in Tab.~\ref{tab:data_summary} the different datasets and their attributes such as their sampling \emph{frequency}, \emph{variates} which determine the number of channels in each example, the total number of \emph{timesteps} in each dataset, the different \emph{horizon} lengths used for forecasting, and lastly the \emph{lookback} period which is the input length used for the prediction.

\begin{table*}[!h]
\centering
\label{tab:data_summary}
{\renewcommand{\arraystretch}{1.5} 
\resizebox{ 0.58 \textwidth}{!}{
\begin{tabular}{ |c| c c c c c   |}
 \hline
 \multicolumn{6}{|c|}{Dataset Summary} \\
 \hline
\shortstack{dataset} & \shortstack{frequency} & \shortstack{variates} & \shortstack{total timesteps} & \shortstack{horizon} & \shortstack{lookback period} \\
 \hline
 \hline
ETTm2 & 15 minutes & 7 & 69,680 &  96, 192, 336, 720 & 96  \\
Electricity & hourly & 321 &  26,304 &  96, 192, 336, 720 & 96 \\
Exchange & daily & 8 & 7,588 &  96, 192, 336, 720 & 96 \\
Traffic & hourly & 862 & 17,544 &  96, 192, 336, 720 & 96  \\
Weather & 10 minutes & 21 & 52,696 &  96, 192, 336, 720 & 96  \\
ILI & weekly & 7 & 966 &  24, 36, 48, 60 & 36  \\

 \hline
\end{tabular}}}
\end{table*}

\subsection{Implementation details}

\paragraph{Baselines.} We train all models based on the implementation and architecture details as they appear in \citep{oreshkin2019n} for N-BEATS and \citep{zhou2021informer, wu2021autoformer,zhou2022fedformer} for the Transformer-based models. The model weights are optimized with respect to the mean squared error (MSE) using the ADAM optimizer \citep{kingma2014adam} with an initial learning rate of $10^{-3}$ for N-BEATS and $10^{-4}$ for Transformer-based models. The maximum number of epochs is set to $10$ allowing early-stopping with a patience parameter of $3$. The reported baseline results are obtained using our environment and hardware, and they may slightly differ from the reported values for the respective methods. Every experiment is run on three different seed numbers, and the results are averaged over the runs. The Pytorch library \citep{paszke2019pytorch} is used for all model implementations, and executed with NVIDIA GeForce RTX 3090 24GB.

\begin{table*}[!h]
\centering
    \caption{Multivariate long-term time-series forecasting results on six datasets in comparison to five baseline models. Low MSE and MAE values are better, and high relative improvement MSE$\%$ and MAE$\%$ scores are better. Boldface text highlights the best performing models.}
    
    \label{tab:multi_results}
   \scalebox{0.75}{
    \begin{tabular}{ll|rr|rr|rr|rr|rrp{1cm}p{1cm}}
    \toprule
        & & \multicolumn{2}{c|}{Informer} & \multicolumn{2}{c|}{Autoformer} & \multicolumn{2}{c|}{FEDformer-w} & \multicolumn{2}{c|}{FEDformer-f} &  \multicolumn{4}{c}{TSAA} \\
        & &      MSE &    MAE &        MSE &    MAE &         MSE &    MAE &         MSE &    MAE &            MSE$\downarrow$ &    MAE$\downarrow$ & MSE$\%$$\uparrow$  & MAE$\%$$\uparrow$  \\
    \midrule
    \multirow{4}{*}{\rotatebox[origin=c]{90}{ETTm2}} & 96  &    0.545 &  0.588 &      0.231 &  0.310 &       0.205 &  0.290 &       0.189 &  0.282 &       \textbf{0.187} &  \textbf{0.274} &      \textbf{1.058} &      \textbf{2.837} \\
        & 192 &    1.054 &  0.808 &      0.289 &  0.346 &       0.270 &  0.329 &       0.258 &  0.326 &       \textbf{0.255} &  \textbf{0.314} &      \textbf{1.163} &      \textbf{3.681} \\
        & 336 &    1.523 &  0.948 &      0.341 &  0.375 &       0.328 &  0.364 &       0.323 &  0.363 &         \textbf{0.304} &  \textbf{0.350} &      \textbf{5.882} &      \textbf{3.581} \\
        & 720 &    3.878 &  1.474 &      0.444 &  0.434 &       0.433 &  0.425 &       0.425 &  0.421 &         \textbf{0.398} &  \textbf{0.403} &      \textbf{6.353} &      \textbf{4.276} \\
    \cmidrule(lr){1-14}
    \multirow{4}{*}{\rotatebox[origin=c]{90}{Electricity}} & 96  &    0.336 &  0.416 &      0.200 &  0.316 &       0.196 &  0.310 &       0.185 &  0.300 &       \textbf{0.183} &  \textbf{0.297} &      \textbf{1.081} &      \textbf{1.000} \\
        & 192 &    0.360 &  0.441 &      0.217 &  0.326 &       0.199 &  0.310 &       0.201 &  0.316 &        \textbf{0.195} &  \textbf{0.309} &      \textbf{2.010} &      \textbf{0.323} \\
        & 336 &    0.356 &  0.439 &      0.258 &  0.356 &       0.217 &  0.334 &       0.214 &  0.329 &        \textbf{0.208} &  \textbf{0.323} &      \textbf{2.804} &      \textbf{1.824} \\
        & 720 &    0.386 &  0.452 &      0.261 &  0.363 &       0.248 &  0.357 &       0.246 &  0.353 &        \textbf{0.238} &  \textbf{0.348} &      \textbf{3.252} &      \textbf{1.416} \\
    \cmidrule(lr){1-14}
    \multirow{4}{*}{\rotatebox[origin=c]{90}{Exchange}} & 96  &    1.029 &  0.809 &      0.150 &  0.281 &       0.151 &  0.282 &       \textbf{0.142} &  \textbf{0.271} &         0.143 &  0.272 &     -0.704 &     -0.369 \\
        & 192 &    1.155 &  0.867 &      0.318 &  0.409 &       0.284 &  0.391 &       0.278 &  0.383 &         \textbf{0.270} &  \textbf{0.378} &      \textbf{2.878} &      \textbf{1.305} \\
        & 336 &    1.589 &  1.011 &      0.713 &  0.616 &       \textbf{0.442} &  \textbf{0.493} &       0.450 &  0.497 &         0.459 &  0.504 &     -3.846 &     -2.231 \\
        & 720 &    3.011 &  1.431 &      1.246 &  0.872 &       1.227 &  0.868 &       \textbf{1.181} &  \textbf{0.841} &    1.213 &  0.842 &     -2.710 &     -0.119 \\
    \cmidrule(lr){1-14}
    \multirow{4}{*}{\rotatebox[origin=c]{90}{Traffic}} & 96  &    0.744 &  0.420 &      0.615 &  0.384 &       0.584 &  0.368 &       0.577 &  0.361 &     \textbf{0.565} &  \textbf{0.352} &      \textbf{2.080} &      \textbf{2.493} \\
        & 192 &    0.753 &  0.426 &      0.670 &  0.421 &       0.596 &  0.375 &       0.610 &  0.379 &        \textbf{0.571} &  \textbf{0.351} &      \textbf{4.195} &      \textbf{6.400} \\
        & 336 &    0.876 &  0.495 &      0.635 &  0.392 &       0.590 &  0.365 &       0.623 &  0.385 &        \textbf{0.584} &  \textbf{0.359} &      \textbf{1.017} &      \textbf{1.644} \\
        & 720 &    1.011 &  0.578 &      0.658 &  0.402 &       0.613 &  0.375 &       0.632 &  0.388 &        \textbf{0.607} &  \textbf{0.368} &      \textbf{0.979} &      \textbf{1.867} \\
    \cmidrule(lr){1-14}
    \multirow{4}{*}{\rotatebox[origin=c]{90}{Weather}} & 96  &    0.315 &  0.382 &      0.259 &  0.332 &       0.269 &  0.347 &       0.236 &  0.316 &      \textbf{0.180} &  \textbf{0.256} &     \textbf{23.729} &     \textbf{18.987} \\
        & 192 &    0.428 &  0.449 &      0.298 &  0.356 &       0.357 &  0.412 &       0.273 &  0.333 &      \textbf{0.252} &  \textbf{0.311} &      \textbf{7.692} &      \textbf{6.607} \\
        & 336 &    0.620 &  0.554 &      0.357 &  0.394 &       0.422 &  0.456 &       0.332 &  0.371 &        \textbf{0.296} &  \textbf{0.355} &     \textbf{10.843} &      \textbf{4.313} \\
        & 720 &    0.975 &  0.722 &      0.422 &  0.431 &       0.629 &  0.570 &       0.408 &  0.418 &     \textbf{0.382} &  \textbf{0.395} &      \textbf{6.373} &      \textbf{5.502} \\
    \cmidrule(lr){1-14}
    \multirow{4}{*}{\rotatebox[origin=c]{90}{ILI}} & 24  &    5.349 &  1.582 &      3.549 &  1.305 &       \textbf{2.752} &  1.125 &       3.268 &  1.257 &    2.760 &  \textbf{1.123} &     -0.291 &      \textbf{0.178} \\
        & 36  &    5.203 &  1.572 &      2.834 &  1.094 &       \textbf{2.318} &  \textbf{0.980} &       2.648 &  1.068 &       2.362 &  0.984 &     -1.898 &     -0.408 \\
        & 48  &    5.286 &  1.594 &      2.889 &  1.122 &       2.328 &  1.006 &       2.615 &  1.072 &    \textbf{2.264} &  \textbf{0.988} &      \textbf{2.749} &      \textbf{1.789} \\
        & 60  &    5.419 &  1.620 &      2.818 &  1.118 &       2.574 &  1.081 &       2.866 &  1.158 &     \textbf{2.520} &  \textbf{1.062} &      \textbf{2.098} &      \textbf{1.758} \\
    \bottomrule
    \end{tabular}}
\end{table*}

\paragraph{Method.} We use Optuna~\citep{akiba2019optuna} for the implementations of TPE and ASHA. The number of trials $T_\text{max}$ is set to $100$. For TPE, In order to guarantee aggressive exploration at the beginning, we run the first  $ 30\% $ of trials with random search. For ASHA, $r$ and $\eta$ are set globally to $1$ and $3$ respectively. The maximum resource parameter $R$, representing the epochs, is set differently for each experiment, due to the baseline's early-stopping.

After the augmentation policy search is finalized, a maximum of $k$ best policies are selected to obtain $p_{\theta^*}$, where $k=3$, and the final model is \emph{fine-tuned} with $\theta^* \sim p_{\theta^*}$ using the shared weights $\omega_\text{share}$. We opt to fine-tune the model and not re-train from random weights so that the final model training matches our optimization process as close as possible. Indeed, \cite{cubuk2020randaugment} discuss the potential differences between the final model behavior in comparison to the performance of the intermediate proxy tasks, i.e., the models obtained during optimization. As the similarity in performance between these models and the final model is not guaranteed, a natural choice is to similarly train the proxy tasks and the final model, as we propose.

\paragraph{Augmentations.}  Each transformation includes a different increasing or decreasing magnitude range which are all mapped to the range $[0,1]$. This way, $m=0$ implies the identity and $m=1$ is the maximum scale. To eliminate cases of the identity being repeatedly chosen, we replace the lower bound in the range with an $\epsilon > 0$ such that for all transformations in the search space only $m > 0$ is possible.
The transformations \textit{Trend scale} and \textit{Seasonality scale} require computing the seasonality and trend components; we pre-compute these factors using the decomposition in STL \citep{cleveland1990stl} and treat it as part of the input data. Each augmentation is applied before the input is fed to the model, namely, on the input $x$ and the target $y$ of the train data batches.

\subsection{Main results}

In our experiments, we employ a similar setup to \citep{ wu2021autoformer,zhou2022fedformer}, where the input length is $96$ and the evaluated forecast horizon corresponds to $96, 192, 336$, or $720$. For ILI, we use input length $36$ and horizons $24, 36, 48, 60$. For a fair comparison, we re-produce all baseline results on our system, and the augmentations are applied on the same generated batches as the baseline. Our main results are summarized in Tab.~\ref{tab:multi_results} and Tab.~\ref{tab:uni_results} including all the baseline results and TSAA. For TSAA, we include the best performing model trained on all baseline architectures. The full results for every architecture with and without TSAA are provided in the appendix spanning tables \ref{tab:informer_mul}-\ref{tab:nbeatsg_uni}. We detail the mean absolute error (MAE) and mean squared error (MSE)~\citep{oreshkin2019n}. Lower values are better, and boldface text highlights the best performing model for each dataset and metric. For TSAA, we also include the \emph{relative improvement} percentage, i.e., $100 \cdot (e_b - e_n) / e_b$, where $e_b$ is the best baseline error and $e_n$ is our result. We denote by MSE$\%$ and MAE$\%$ the relative improvement of MSE and MAE, respectively. A higher improvement is better. 

\begin{table*}[!t]
\centering
    \caption{Univariate long-term time-series forecasting results on five datasets in comparison to five baseline models. Low MSE and MAE values are better, and high relative improvement MSE$\%$ and MAE$\%$ scores are better. Boldface text highlights the best performing models.}
    \label{tab:uni_results}
   \scalebox{0.75}{
    \begin{tabular}{ll|rr|rr|rr|rr|rr|rrp{1.2cm}p{1.2cm}}
    \toprule
        & & \multicolumn{2}{c|}{Informer} & \multicolumn{2}{c|}{Autoformer} & \multicolumn{2}{c|}{FEDformer-f}& \multicolumn{2}{c|}{N-BEATS-I} &  \multicolumn{2}{c|}{N-BEATS-G} & \multicolumn{4}{c}{TSAA} \\
        &  &            MSE &    MAE &         MSE &    MAE &         MSE &    MAE &         MSE &    MAE &    MSE &     MAE & MSE$\downarrow$  & MAE$\downarrow$ & MSE$\%$$\uparrow$  & MAE$\%$$\uparrow$ \\
    \midrule
    \multirow{4}{*}{\rotatebox[origin=c]{90}{ETTm2}} & 96  &    0.085 &  0.225 &      0.123 &  0.270 &       \textbf{0.068} &  0.198 &    0.080 &  0.213 &    0.080 &  0.210 &         \textbf{0.068} &  \textbf{0.192} &      \textbf{0.000} &      \textbf{3.030} \\
            & 192 &    0.130 &  0.282 &      0.141 &  0.289 &       \textbf{0.096} &  0.238 &    0.103 &  0.240 &    0.110 &  0.250 &        \textbf{0.096} &  \textbf{0.237} &      \textbf{0.000} &      \textbf{0.420} \\
            & 336 &    0.161 &  0.314 &      0.170 &  0.319 &      \textbf{ 0.138} &  \textbf{0.286} &    0.162 &  0.312 &    0.172 &  0.320 &       0.139 &  0.290 &     -0.725 &     -1.399 \\
            & 720 &    0.221 &  0.373 &      0.206 &  0.353 &       0.189 &  \textbf{0.335} &    0.199 &  0.347 &    0.201 &  0.353 &         \textbf{0.187} &  0.336 &     \textbf{ 1.058} &     -0.299 \\
            \cmidrule(lr){1-16}
    \multirow{4}{*}{\rotatebox[origin=c]{90}{Electricity}} & 96  &    0.261 &  0.367 &      0.454 &  0.508 &       \textbf{0.244} &  0.364 &    0.326 &  0.402 &    0.324 &  0.397 &        \textbf{0.244} &  \textbf{0.354} &      \textbf{0.000} &      \textbf{2.747} \\
            & 192 &    0.285 &  0.386 &      0.511 &  0.532 &       \textbf{0.276} &  0.382 &    0.350 &  0.417 &    0.363 &  0.420 &      0.277 &  \textbf{0.368} &     -0.362 &      \textbf{3.665} \\
            & 336 &    0.324 &  0.417 &      0.739 &  0.651 &       0.347 &  0.432 &    0.393 &  0.440 &    0.392 &  0.443 &       \textbf{0.310} &  \textbf{0.394} &      \textbf{4.321} &      \textbf{5.516} \\
            & 720 &    0.632 &  0.612 &      0.673 &  0.610 &       0.408 &  0.473 &    0.458 &  0.490 &    0.489 &  0.502 &        \textbf{0.378} &  \textbf{0.447} &      \textbf{7.353} &      \textbf{5.497} \\
            \cmidrule(lr){1-16}
    \multirow{4}{*}{\rotatebox[origin=c]{90}{Exchange}} & 96  &    0.490 &  0.554 &      0.149 &  0.308 &       0.133 &  0.284 &    0.210 &  0.344 &    0.223 &  0.351 &    \textbf{0.093} &  \textbf{0.236} &     \textbf{30.075} &     \textbf{16.901} \\
            & 192 &    0.790 &  0.721 &      0.290 &  0.415 &       0.292 &  0.419 &    1.130 &  0.840 &    0.783 &  0.675 &    \textbf{0.215} &  \textbf{0.352} &     \textbf{25.862} &     \textbf{15.181} \\
            & 336 &    2.146 &  1.223 &      0.708 &  0.662 &      \textbf{0.477} &  \textbf{0.532} &    1.587 &  1.047 &    2.622 &  1.266 &     0.532 &  0.572 &    -11.530 &     -7.519 \\
            & 720 &    1.447 &  1.008 &      1.324 &  0.892 &       1.304 &  0.882 &    0.870 &  0.747 &    2.588 &  1.303 &     \textbf{0.527} &  \textbf{0.594} &     \textbf{39.425} &     \textbf{20.482} \\
            \cmidrule(lr){1-16}
    \multirow{4}{*}{\rotatebox[origin=c]{90}{Traffic}} & 96  &    0.262 &  0.348 &      0.266 &  0.372 &       0.210 &  0.318 &    0.181 &  0.268 &    0.159 &  0.240 &       \textbf{0.158} &  \textbf{0.239} &      \textbf{0.629} &      \textbf{0.417} \\
            & 192 &    0.294 &  0.376 &      0.272 &  0.379 &       0.206 &  0.311 &    0.177 &  0.263 &    0.181 &  0.264 &   \textbf{0.160} &  \textbf{0.243} &      \textbf{9.605} &      \textbf{7.605} \\
            & 336 &    0.308 &  0.390 &      0.261 &  0.374 &       0.217 &  0.322 &    0.180 &  0.271 &    \textbf{0.155} &  \textbf{0.239} &     0.156 &  0.244 &     -0.645 &     -2.092 \\
            & 720 &    0.364 &  0.440 &      0.269 &  0.372 &       0.243 &  0.342 &    0.226 &  0.316 &    0.212 &  0.304 &    \textbf{0.189} &  \textbf{0.279} &     \textbf{10.849} &      \textbf{8.224} \\
            \cmidrule(lr){1-16}
    \multirow{4}{*}{\rotatebox[origin=c]{90}{Weather}} & 96  &    0.005 &  0.048 &      0.009 &  0.078 &       0.009 &  0.073 &    0.003 &  0.044 &    0.003 &  0.043 &     \textbf{0.001} &  \textbf{0.024} &     \textbf{66.667} &     \textbf{44.186} \\
            & 192 &    0.004 &  0.051 &      0.009 &  0.068 &       0.007 &  0.067 &    0.004 &  0.046 &    0.004 &  0.047 &    \textbf{0.001} &  \textbf{0.027} &     \textbf{75.000} &     \textbf{41.304} \\
            & 336 &    0.003 &  0.043 &      0.006 &  0.058 &       0.006 &  0.062 &    0.004 &  0.048 &    0.005 &  0.054 &   \textbf{0.002} &  \textbf{0.035} &     \textbf{33.333} &     \textbf{18.605} \\
            & 720 &    0.004 &  0.049 &      0.007 &  0.063 &       0.006 &  0.060 &    0.004 &  0.049 &    0.004 &  0.048 &     \textbf{0.002} &  \textbf{0.034} &     \textbf{50.000} &     \textbf{29.167} \\
    \bottomrule
    \end{tabular}}
\end{table*} 

\paragraph{Multivariate time-series forecasting results.} Based on the results in Tab.~\ref{tab:multi_results}, we observe that most datasets benefit from automatic augmentation, where in the vast majority of cases, TSAA improves the baseline scores. It is apparent that TSAA yields stronger performance in particular in the long-horizon settings with $\textbf{6.35\%}$ ($0.425 \rightarrow 0.398$) reduction in ETTm2, $\textbf{3.25\%}$ ($0.246 \rightarrow 0.238$) reduction in Electricity, and $\textbf{2.1\%}$ ($2.328 \rightarrow 2.264$) reduction in ILI. One of the more prominent results appears for Weather $96$ and $336$ with reductions in MSE of $\textbf{23.73\%}$ ($0.236 \rightarrow 0.180$) and $\textbf{10.84\%}$ ($0.332 \rightarrow 0.296$), respectively. For the Exchange dataset, TSAA obtains slightly higher errors with respect to the FEDformer-w baseline. Overall, TSAA achieves the \emph{best results} in $39$ error metrics, in comparison to FEDformer-f and FEDformer-w with $4$ and $5$ best models, respectively. 


\paragraph{Univariate time-series forecasting results.} Similar to the multivariate results, most long horizon settings benefit from TSAA. With a $21.74\%$ average reduction across all datasets with a horizon of $720$. Furthermore, the results that stand out the most are the MSE and MAE reductions in Weather, with a $66\%, 75\%, 33, 50\%\%$, and respectively $44.2\%, 41.3\%, 17.6\%, 29.2\%$ performance improvements corresponding to the $96, 192, 336$ ,and $720$ horizons. Further, it is evident in Tabs.~\ref{tab:informer_uni}-\ref{tab:nbeatsg_uni}, that the improvements in the Weather dataset are not limited to a specific baseline architecture. In contrast to the multivariate setting, TSAA achieves significantly better scores on the Exchange dataset with average improvements of $21\%$ and $11.27\%$ for the MSE and MAE metrics. Notably, the results in the univariate case are slightly more involved than the multivariate setting such that that only Weather always benefits from TSAA, whereas the results for other datasets are mixed. Still, TSAA shows a positive advantage over all baseline models. In particular, TSAA obtained the best models for $32$ error metrics, whereas FEDformer-f and N-BEATS-G are better in $9$ and $2$ measures.

\begin{figure*}[t]
    \centering
   \includegraphics[width=.9\linewidth]{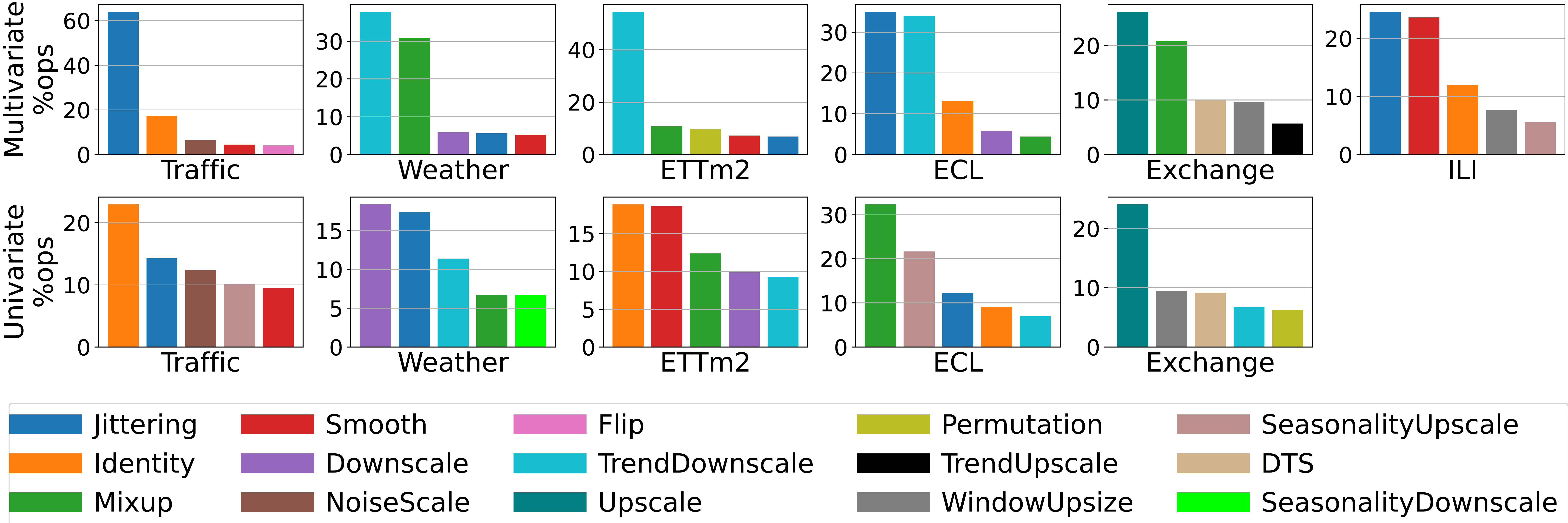}
   \caption{The best five performing transformations per dataset attained with TSAA, measured with the percentage proportion of the selected operations (\%ops). Each colored bar represents a transformation and the y-axis represents the percentage proportion the given transformation accounts for.}
   \label{fig:top_transformations}
\end{figure*}

\begin{figure}[h!]
    \centering
    \includegraphics[width=\linewidth]{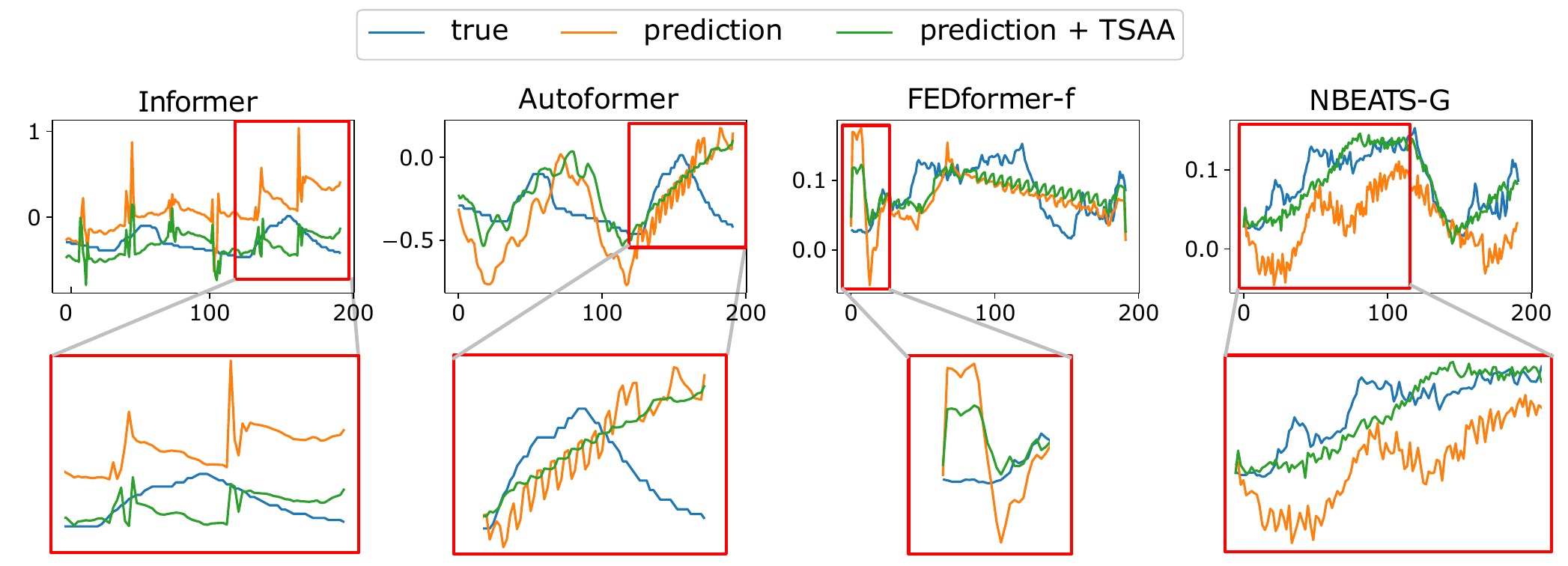} 
    \includegraphics[width=\linewidth, trim={0cm 0cm 0 7.3cm}, clip]{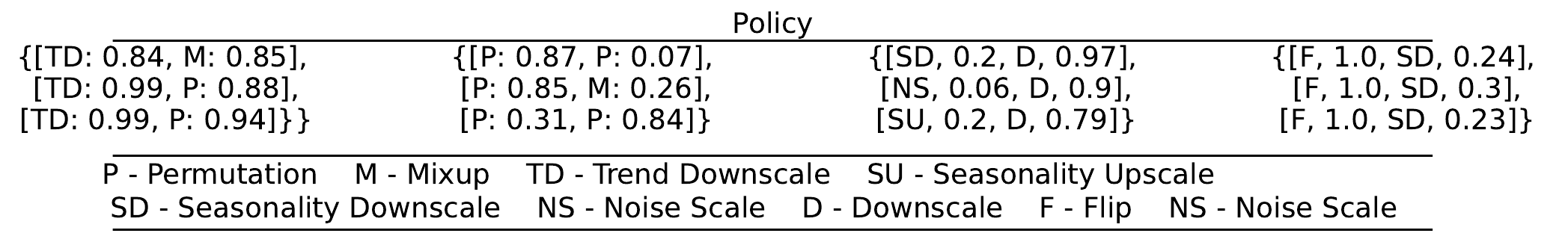} 
    \caption{The ground truth, prediction, and prediction with augmentation attained with TSAA applied to the same forecast target in ETTm2 with Informer, Autoformer multivariate, and Weather with FEDformer-f and NBEATS-G univariate. It is shown that augmentation can assist the different models to achieve more accurate predictions with better alignment, reduce excessive noise, and curtail outliers. The attained policies are given underneath each plot.} \label{fig:prediction_example}
\end{figure}

\paragraph{Policy analysis.} The most noticeable selected transformations are illustrated in Fig.~\ref{fig:top_transformations}. It is evident that the transformations Trend Downscale, Jittering, Mixup, and Smoothing are some of the prominent selections in the overall setup. Trend Downscale accounts for more than $30\%$ of the operations in ETTm2, Weather and Electricity; this may indicate that the deep models tend to overestimate the trend, and thus it requires downscaling. Jittering and Smoothing on the other hand, do not violate time-series characteristics such as trend or seasonality but still promote diversity within the given dataset, where Smoothing is approximately the opposite of Jittering. Notably, Mixup appeared as one of the five most important transformations for four and three datasets in the multivariate and univariate settings, respectively. We believe that Mixup is beneficial to TSF since it samples from a vicinal distribution whose variability is higher than the original train set. We show in Fig.~\ref{fig:prediction_example} the outcome with and without TSAA compared to the ground truth, showing that employing custom policies per signal may significantly improve forecasting.

\section{Ablation and Analysis}
\label{sec:ablation}

\subsection{Parameter selection}

\begin{figure}[!h]
  \centering
  \begin{overpic}[width=0.9\linewidth]{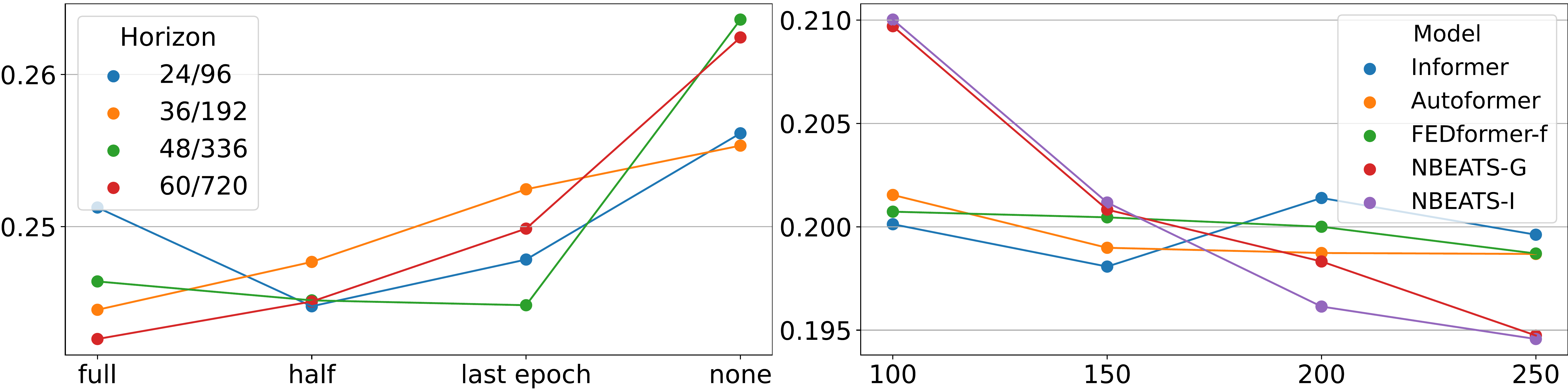} 
        \put(1, 1){A} \put(51.5, 1){B}
  \end{overpic}
  \caption{A) The normalized average performance measures as a function of different $\beta$ values. Our results indicate that $\beta=0.5$ (half) attains the best computational resources to performance gain ratio. B) We plot the normalized average performance measures as a function of different $T_\text{max}$ values. As the number of trials grows, we observe better overall performance.}  
  \label{fig:beta_trials}
\end{figure}

\paragraph{Choice of $\beta$.} In what follows, we motivate our choice for the $\beta$ hyperparameter which dictates for how many epochs we pre-train the baseline architecture to obtain $\omega_\text{share}$. To this end, we investigate the effect of utilizing different values of $\beta$. We consider four different settings: 1) full training with augmentation, i.e., $\beta = 0.0$, 2) half training with augmentation, i.e., $\beta = 0.5$, 3) augmentation applied only in the last epoch, 4) baseline training with no augmentation, i.e., $\beta = 1.0$. We used TSAA on the ILI with respect to N-BEATS-G in the univariate setting, and Informer, Autoformer and FEDformer-f in the multivariate case, as well as on multivariate ETTm2 with Autoformer and FEDformer-f. We plot the averaged results of these architectures in Fig.~\ref{fig:beta_trials}A, showing four colored curves corresponding to the various forecasting horizons $24/96, \  36/192, \  48/336$ and $60/720$ with colors blue, orange, green and red, respectively. The best models are obtained for $\beta=0.0$ and $\beta=0.5$, that is, full- and half-augmented training. Somewhat surprisingly, two of the four best forecasting horizons ($36/192, 48/336$) are obtained for $\beta=0.5$. Overall, the fully augmented model (i.e., $\beta=1.0$) attains a $\textbf{5.1\%}$ average improvement over the baseline, whereas using $\beta=0.5$ yields a $\textbf{5.3\%}$ average improvement. Thus, fully training with augmentation achieves similar performance to half training, while requiring significantly more resources. Indeed, \cite{tian2020improving} employs a similar strategy, and thus, we propose to generate $\omega_\text{share}$ after training for \emph{half} of the active epochs, and we fine-tune the model using optimal augmentation policies.

\textbf{Reduction factor and linked operations.} In Sec.~\ref{sec:background}, we introduced the reduction factor $\eta$ that controls the number of kept runs in ASHA. Additionally, we discuss in Sec.~\ref{sec:method} that every sub-policy is composed of $n$ linked operations of time-series augmentations. Here, we would like to empirically justify our choices for these two hyperparameters. Our ablation study uses the ILI dataset on the $36, 48, 60$ forecasting tasks, with N-BEATS-G for the univariate case, and Informer, Autoformer, and FEDformer-f for the multivariate configuration. We test the values $\eta \in \{2, 3\}$ and $n \in \{1, 2\}$. Every experiment is repeated three times, and we analyze the average results. 

Overall, we propose to use the values $\eta = 3$ and $n = 2$ due to the following observations arising from our experiments. The improvement difference between $\eta = 2$ and $\eta = 3$ is only \textbf{0.12\%} in favor of $\eta=2$, thus suggesting that neither exhibits a statistically-dominant performance advantage. Nevertheless, $\eta = 3$ is resource efficient as it reduces the amount of kept runs $1/\eta$ by $16.67\%$. Moreover, a single operation $n=1$ attains a $\textbf{6.4\%}$ average improvement compared to the baseline, whereas two linked operations $n=2$ yield a $\textbf{7.4\%}$ average improvement.

\paragraph{Convergence of TSAA.} In our experiments, we look for good augmentation policies for $T_\text{max} = 100$ iterations. Here, we explore the effect of this value on the performance of the resulting models. We evaluate our framework on the ILI dataset with the architectures Informer, Autoformer, FEDformer-f, N-BEATS-G and N-BEATS-I using varying values for $T_\text{max} \in \{ 100, 150, 200, 250\}$. Intuitively, greater $T_\text{max}$ values may result in an improved convergence and a better overall performance as the framework can explore and exploit a larger variety of configurations from the search space. Indeed, we show in Fig.~\ref{fig:beta_trials}B the normalized average MSE values obtained for the various tests. We observe an MSE reduction of $\textbf{1\%}$ for the transformer-based models when increasing $T_\text{max}=100$ to $T_\text{max}=250$. The N-BEATS architecture benefited more and achieved a $\textbf{7.25\%}$ reduction. In conclusion, the hyperparameter $T_\text{max}$ presents a natural trade-off to the practitioner: higher $T_\text{max}$ values generally lead to better performance at a higher computational cost, whereas lower values are less demanding computationally but present inferior performance.

\begin{table*}[!h]
 \centering
    \caption{Comparison of automatic augmentation approaches including TSAA, Fast AutoAugment and RandAugment. We denote by $\%$AI the average improvement in percentage.}
    \label{tab:aa_results}
    \scalebox{0.8}{
\begin{tabular}{ll|cc|cc|cc|cc}
\toprule
    &  & \multicolumn{2}{c|}{Best Baseline} & \multicolumn{2}{c|}{Fast AA} & \multicolumn{2}{c|}{RandAugment} & \multicolumn{2}{c}{TSAA} \\
    & Metric &    MSE &    MAE &         MSE &    MAE &           MSE &    MAE &    MSE &    MAE \\
\midrule
\multirow{4}{*}{\rotatebox[origin=c]{90}{ETTm2}} & 96  &  \underline{ 0.189} &  \underline{0.282} & 0.197 &  0.279 &       0.192 &  0.282 &   \textbf{0.187} &  \textbf{0.274} \\
    & 192 & 0.258 &  0.326 & 0.262 &  \underline{0.319} &       \underline{0.257} &  0.323 &           \textbf{0.255} &  \textbf{0.314} \\
    & 336 & 0.323 &  0.363 &  \underline{0.322} &  \underline{0.356} &       0.326 &  0.364 &          \textbf{0.311} &  \textbf{0.350} \\
    & 720 & 0.425 &  0.421 &  \underline{0.415} &  \underline{0.407} &       0.427 &  0.420 &          \textbf{0.406} &  \textbf{0.403} \\ \cmidrule(lr){1-10}
\multirow{4}{*}{\rotatebox[origin=c]{90}{Traffic}} & 96  &  \textbf{0.577} &  \textbf{0.361} & 0.655 &  0.410 &       \underline{0.590} &  0.376 &      \textbf{0.577} &  \underline{0.362} \\
    & 192 & \underline{0.610} &  \underline{0.379} & 0.652 &  0.408 &       0.622 &  0.390 &           \textbf{0.601} &  \textbf{0.371} \\
    & 336 &  0.623 &  0.385 &  0.674 &  0.421 &       0.626 &  0.392 &         \textbf{0.619} &  \textbf{0.383} \\
    & 720 & \textbf{0.632} &  \textbf{0.388} & 0.705 &  0.427 &       0.643 &  0.396 &           \textbf{0.632} &  \textbf{0.388} \\ \cmidrule(lr){1-10}
\multirow{4}{*}{\rotatebox[origin=c]{90}{Weather}} & 96  &  0.236 &  0.316 & \textbf{0.191} &  \textbf{0.252} &       \underline{0.203} &  \underline{0.275} &         0.207 &  0.285 \\
    & 192 & 0.273 &  0.333 &  \textbf{0.240} &  \textbf{0.290} &       0.267 &  0.332 &          \underline{0.252} &  \underline{0.311} \\
    & 336 &  0.332 &  0.371 &  \textbf{0.290} &  \textbf{0.321} &       0.328 &  0.364 &     \underline{0.313} &  \underline{0.355} \\
    & 720 & 0.408 &  0.418 &  \textbf{0.363} &  \textbf{0.367} &       0.398 &  0.413 &          \underline{0.382} &  \underline{0.395} \\ \cmidrule(lr){1-10}
\multirow{4}{*}{\rotatebox[origin=c]{90}{ILI}} & 24  & 3.268 &  1.257 & 4.671 &  1.603 &       \underline{3.160} &  \underline{1.234} &           \textbf{3.150} &  \textbf{1.219} \\
    & 36  & 2.648 &  1.068 &   3.835 &  1.375 &       \textbf{2.457} &  \textbf{1.019} &          \underline{2.578} &  \underline{1.049} \\
    & 48  &  2.615 &  1.072 & 3.694 &  1.359 &      \textbf{ 2.558} &  \textbf{1.045} &       \underline{2.609} &  \underline{1.069} \\
    & 60  & 2.866 &  1.158 & 3.855 &  1.410 &       \textbf{2.775} &  \textbf{1.108} &          \underline{2.805} &  \underline{1.140} \\ \cmidrule(lr){1-10} \cmidrule(lr){1-10}

 \multicolumn{2}{c|}{$\%$AI}  & 0.0 &  0.0 & -9.5 &  -4.88 &      1.67 &  1.22 &   \textbf{3.33} &  \textbf{3.1} \\
    \bottomrule
    \end{tabular}}
\end{table*} 

\subsection{AutoAugment Method Comparison}
\label{app:aa_comparison}

We compare TSAA to other efficient AutoAugment methods via an experiment that shows that both Fast AutoAugment~\citep{lim2019fast} and RandAugment~\citep{cubuk2020randaugment} are \emph{inconsistent} and thus inferior to TSAA. In this experiment, we tested the performance of deploying Fast AutoAugment and RandAugment with the same search space, with the exception of discretized magnitude ranges so RandAugment falls in line with the original method. The given methods were tested on Autoformer and FEDformer-f multivariate together with datasets: ETTm2, Traffic, Weather, and ILI. The results are provided in Tab. \ref{tab:aa_results}. For Fast AutoAugment we set $K=3$ folds which control the number of subsets, each subset explores and exploits $100$ augmentation trials. For RandAugment, we discretize the magnitude range to $8$ bins and utilize the partial train scheme with the same $\beta$ as in TSAA, to allow RandAugment to benefit from the same approach. While it is shown that Fast AutoAugment and RandAugment are superior on Weather and ILI respectively, they attain inferior results on the other datasets. Nevertheless, TSAA is shown to be most effective on Traffic and ETTm2 and second-best on ILI and Weather. Additionally, TSAA maintains a consistent improvement across all datasets providing a \textbf{$3.33 \%$} average MSE reduction as opposed to RandAugment which offers approximately half of that, or Fast AutoAugment with a negative average MSE reduction. In conclusion, TSAA is more consistent with better overall performance when compared to prominent AutoAugment methods in the time-series domain.

\section{Conclusion}

In this work, we study the task of data augmentation in the setting of time-series forecasting. While recent approaches based on automatic augmentation achieved state-of-the-art results in image classification tasks, problems involving arbitrary time-series information received less attention. Thus, we propose a novel time-series automatic augmentation (TSAA) method that relaxes a difficult bilevel optimization. In practice, our framework performs a partial training of the baseline architecture, followed by an iterative split process. Our iterations alternate between finding the best DA policy for a given set of model weights, to fine-tuning the model based on a specific policy. In comparison to several strong methods on multiple univariate and multivariate benchmarks, our framework improves the baseline results in the majority of prediction settings.

In the future, we would like to explore better ways for relaxing the bilevel optimization, allowing to train an end-to-end model~\citep{li2020dada, zheng2022deep}. Further, we believe that our approach would benefit from stronger time-series augmentation transformations. Thus, one possible direction forward is to incorporate learnable DA modules, similar in spirit to filters of convolutional models.

\subsubsection*{Acknowledgments}
This research was partially supported by an ISF grant 668/21, an ISF equipment grant, and by the Israeli Council for Higher Education (CHE) via the Data Science Research Center, Ben-Gurion University of the Negev, Israel.

\clearpage
\bibliography{main}
\bibliographystyle{tmlr}

\clearpage

\appendix
\section{Appendix}
In what follows, we present additional details the transformations used in our method (App.~\ref{app:ts_da}),  hyperparameter values (App.~\ref{app:hyperparam}), complexity analysis of TSAA (App.~\ref{app:complexity}), and lastly, we provide extended result tables for every baseline alongside TSAA (App.~\ref{app:ex_results}).

\section{Time-series transformations}
\label{app:ts_da}

\begin{figure}[t]
    \centering
   \includegraphics[width=0.82\linewidth]{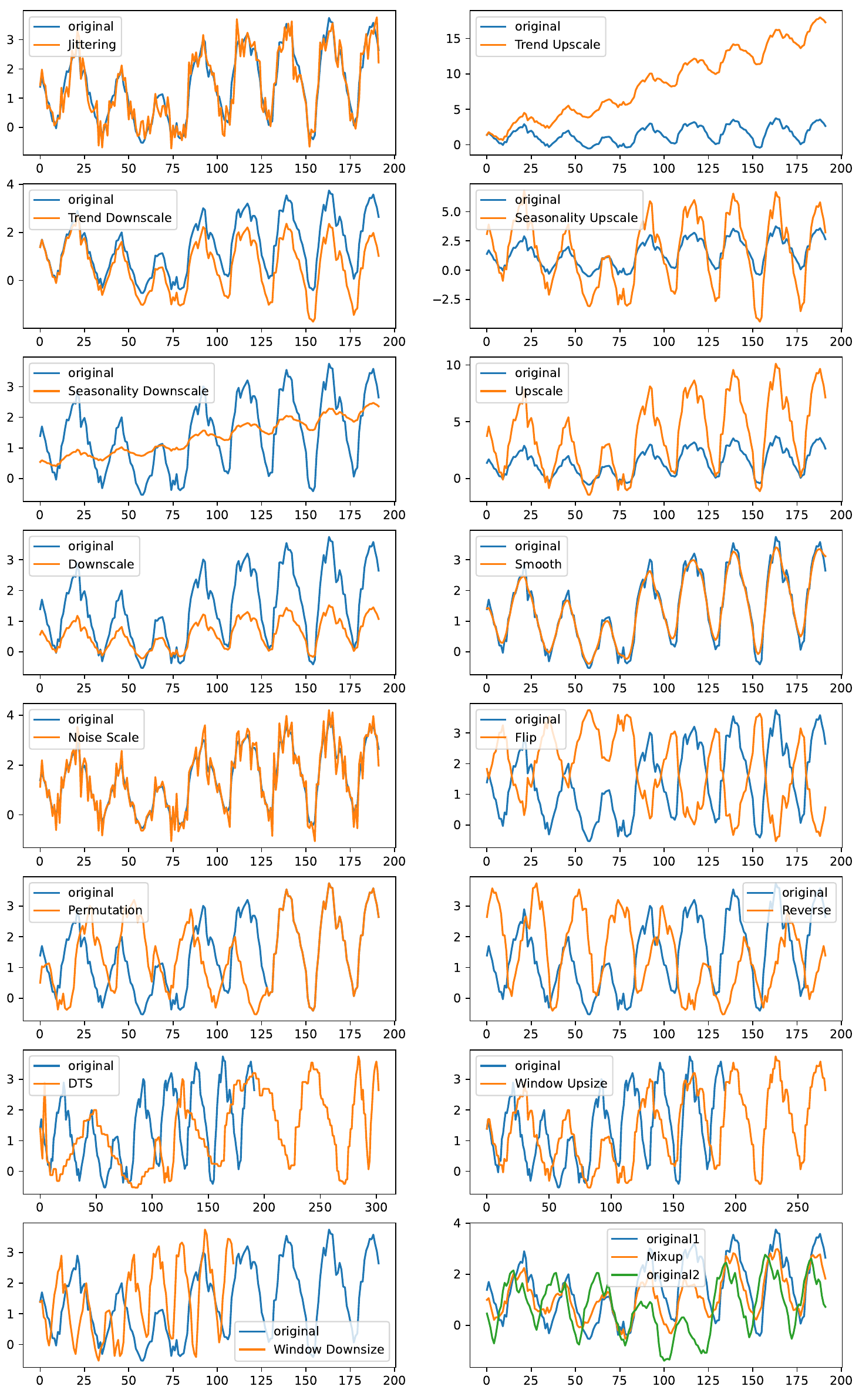}
   \caption{We demonstrate the effect of different transformations when applied to the same example from the Electricity dataset. Blue and orange represent the original signal and its transformed version, respectively.}
   \label{fig:transformations}
\end{figure}

In this section, we offer a detailed description of the different time-series transformations, which can be found in Tab.~\ref{tab:ts_da} and depicted in Fig.~\ref{fig:transformations} with $m = 0.85$ compared to the original signal. For each of the transformations \textit{Trend scale}, \textit{Scale}, \textit{Seasonality scale}, \textit{Window warping}, we use two separate and independent transformations to demonstrate an increase or decrease of the given effect.

\begin{table}[h!]
\caption{Our search space is composed of the following time-series transformations and their associated magnitude range.}
\label{tab:ts_da}

\begin{center}
{\renewcommand{\arraystretch}{1.5} 
\resizebox{ 0.8\textwidth}{!}{
\begin{tabular}{lp{8cm}p{2.9cm}r}

\toprule
Transformation & Description & Range of magnitudes  \\
\midrule
Jittering    & Adds white noise with $\sigma$ controlled by $m$ \citep{um2017data}. *  & [0,0.1]  \\
Trend scale & Multiplies the trend component by $m$. *  & [1,10], [0,1]\\
Scale    & Multiplies the entire series by $m$ \citep{um2017data}. & [1,3], [0.3,1] \\
Seasonality scale & Multiplies the seasonality component by $m$. & [1,3], [0,1]  \\
Smooth & Performs low-pass filtering with a convolution kernel, where $m$ controls the kernel size.& [0,11] \\
Noise scale  & Performs high-pass filtering with a second-order convolution kernel to extract the difference, which is then multiplied by $m$ and added back to the original series. & [0,1] \\
Permutation  & Exchanges two non-overlapping time intervals, such that the interval size is controlled by $m$. \citep{um2017data}. & [0,0.3]\\
Dynamic time stretching  & Manipulates the length of different non-overlapping time intervals, where $m$ controls the manipulation magnitude. \citep{nguyen2020improving}.& [1,5] \\
Window warping  &  Manipulates the length of the entire window \citep{um2017data}.& [1,1.5], [0.5,1]  \\
Mixup  & Linearly interpolates between two series, $m$ controls the contribution of each series \citep{zhang2017mixup}. & [0,0.5] \\
Identity    & Returns the original series.& None \\
Flip    & Flip the series relative to the value location by multiplying by $(-1)$ \citep{iwana2021empirical}. *  & \{0,1\} \\
Reverse    & Change the relative location of the time steps to span from end to start.& \{0,1\} \\
\bottomrule
\end{tabular}}}
\end{center}
\vskip -0.1in

\vspace{1ex}

{\raggedright * marks a transformation implemented with min-max scaling to ensure equal relative changes. }\par
\end{table}

\clearpage

\clearpage

\section{TSAA hyperparameters}
\label{app:hyperparam}

We detail in Tab.~\ref{tab:hyperparam_tsaa_ra} the hyperparameter values we used in the evaluation of TSAA.

\begin{table*}[!ht]
 \centering
     \caption{Hyperparameter values used in the evaluation of TSAA.}
    \label{tab:hyperparam_tsaa_ra}
    \scalebox{0.9}{
\begin{tabular}{ |c c c c c c c c c |}
 \hline
 \multicolumn{9}{|c|}{Model Parameters - TSAA} \\
 \hline
\shortstack{$T_{max}$}& \shortstack{exploration \\ trials fraction} & \shortstack{ $m$ range } & \shortstack{$n$} & \shortstack{$\eta$}& \shortstack{$r$} & \shortstack{resource type} & \shortstack{$\beta$} & \shortstack{$k$}  \\
 \hline
 \hline
 100 & 0.3 & (0,1] & 2 &  3 & 1 & epoch & 0.5 & 3  \\
 \hline
    \end{tabular}}
\end{table*}



\section{Complexity}
\label{app:complexity}

A straightforward upper bound of our method is given by $\mathcal{O}((1-\beta) K T_{max})$ evaluated in epochs, where $K$, $(1-\beta)$, and $T_{max}$ correspond to the number of active epochs the model trains for, the fraction of $K$ to be used for fine-tuning, and the maximum number of trials, respectively. However, for TSAA to practically reach such an upper bound when $K>2$  would require each trial to outperform the preceding trials, thus avoiding being pruned by ASHA. This setup is very unlikely for the following reasons: (1) a fraction size $0.3$ of $T_{max}$ of starting trials are manually dedicated to random search to promote aggressive exploration at the start. (2) To the best of our knowledge Bayesian Optimization does not guarantee monotonic improvement and inherently promotes exploration \citep{bergstra2011algorithms}. (3) $\mathcal{L}_\text{val}(\theta,w^*)$ is not promised to be convex with respect to $\theta$, making it even more difficult to attain a monotonic improvement. In our empirical evaluations, we have observed different operations with different transformations being selected, as opposed to a single policy being repeatedly selected, which strengthens our claim. Further, we would like to share a solid example from our empirical experiments with FEDformrer-w applied to Electricity with a horizon of $336$. This setup is considered resource expensive as one epoch may take as long as $20$ minutes on a single RTX3090. In the given setup, $K$ was set to $8$ and $\beta=0.5$ as defined in the hyperparameter table Tab.~\ref{tab:hyperparam_tsaa_ra}. Therefore, according to the bound mentioned above, the total number of epochs would be $400$ epochs. However, the use of ASHA with Bayesian optimization reduced it to $170$, proving the computational impact. For comparison, applying a naive approach without ASHA and $\beta$ will result in a total number of $800$ training epochs.

\section{Extended results}

In this section, we provide the performance of TSAA for each individual model featured in Tab.~\ref{tab:multi_results} and Tab.~\ref{tab:uni_results} in the main text. We can observe from the multivariate results that Informer benefits the most from TSAA, with the $720$ horizon in particular. On the other end, TSAA struggles to achieve significant improvements in FEDformer-f, especially for the Exchange dataset. In the univariate setting, unlike the multivariate setting, the models NBEATS-G and NBEATS-I gain large improvement rates in Exchange thanks to TSAA. Overall, we can conclude that all models and datasets exhibit a degree of improvement with the support of TSAA.

\clearpage

\subsection{TSF results}
\label{app:ex_results}

Tabs.~\ref{tab:informer_mul}, \ref{tab:autoformer_mul}, \ref{tab:fedformerf_mul}, \ref{tab:fedformerw_mul}, \ref{tab:informer_uni}, \ref{tab:autoformer_uni}, \ref{tab:fedformerf_uni}, \ref{tab:nbeatsi_uni}, \ref{tab:nbeatsg_uni}  detail the extended results for every baseline architecture and dataset we considered in the main text.

\begin{table*}[h!]
\begin{minipage}[c]{0.5\textwidth}
\centering
\caption{Informer multivariate}
\label{tab:informer_mul}
\scalebox{0.67}{

\begin{tabular}{ll|rr|rrrr}
    \toprule
        & &  \multicolumn{2}{c|}{Informer} & \multicolumn{4}{c}{TSAA} \\
        & &           MSE &    MAE &    MSE$\downarrow$ &    MAE$\downarrow$ & MSE$\%$$\uparrow$  & MAE$\%$$\uparrow$  \\
    \midrule
    \multirow{4}{*}{\rotatebox[origin=c]{90}{ETTm2}} & 96  &    0.545 &  0.588 &  \textbf{0.224} &  \textbf{0.321} &     \textbf{58.899} &    \textbf{45.408} \\
        & 192 &    1.054 &  0.808 &  \textbf{0.270} &  \textbf{0.355} &     \textbf{74.383} &     \textbf{56.064} \\
        & 336 &    1.523 &  0.948 &  \textbf{0.304} &  \textbf{0.374} &     \textbf{80.039} &     \textbf{60.549} \\
        & 720 &    3.878 &  1.474 &  \textbf{0.398} &  \textbf{0.435} &     \textbf{89.737} &     \textbf{70.488} \\  \cmidrule(lr){1-8}
    \multirow{4}{*}{\rotatebox[origin=c]{90}{Electricity}}  & 96  &    0.336 &  0.416 &  \textbf{0.324} &  \textbf{0.407} &      \textbf{3.571} &      \textbf{2.163} \\
        & 192 &    0.360 &  0.441 &  \textbf{0.336} &  \textbf{0.419} &      \textbf{6.667} &      \textbf{4.989} \\
\bottomrule
\end{tabular}}

\end{minipage}
\begin{minipage}[c]{0.5\textwidth}
\centering
\caption{Autoformer multivariate}
\label{tab:autoformer_mul}
    \scalebox{0.67}{
    \begin{tabular}{ll|rr|rrrr}
    \toprule
        & &  \multicolumn{2}{c|}{Autoformer} & \multicolumn{4}{c}{TSAA} \\
        & &           MSE &    MAE &    MSE$\downarrow$ &    MAE$\downarrow$ & MSE$\%$$\uparrow$  & MAE$\%$$\uparrow$  \\
    \midrule
    \multirow{4}{*}{\rotatebox[origin=c]{90}{ETTm2}} & 96  &      0.231 &  0.310 &  \textbf{0.211} &  \textbf{0.293} &      \textbf{8.658} &      \textbf{5.484} \\
        & 192 &      0.289 &  0.346 &  \textbf{0.269} &  \textbf{0.327} &      \textbf{6.920} &      \textbf{5.491} \\
        & 336 &      0.341 &  0.375 &  \textbf{0.322} &  \textbf{0.359} &      \textbf{5.572} &      \textbf{4.267} \\
        & 720 &      0.444 &  0.434 &  \textbf{0.410} &  \textbf{0.407} &      \textbf{7.658} &      \textbf{6.221} \\ \cmidrule(lr){1-8}
    \multirow{4}{*}{\rotatebox[origin=c]{90}{Electricity}} & 96  &      0.200 &  0.316 &  \textbf{0.188} &  \textbf{0.302} &      \textbf{6.000} &      \textbf{4.430} \\
        & 192 &      \textbf{0.217} &  \textbf{0.326} &  0.221 &  0.328 &     -1.843 &     -0.613 \\
\bottomrule
\end{tabular}}

\end{minipage}

\begin{minipage}[c]{0.5\textwidth}
\centering
\caption{FEDformer-f multivariate}
\label{tab:fedformerf_mul}
\scalebox{0.67}{

\begin{tabular}{ll|rr|rrrr}
    \toprule
        & &  \multicolumn{2}{c|}{FEDformer-f} & \multicolumn{4}{c}{TSAA} \\
        & &           MSE &    MAE &    MSE$\downarrow$ &    MAE$\downarrow$ & MSE$\%$$\uparrow$  & MAE$\%$$\uparrow$  \\
    \midrule
    \multirow{4}{*}{\rotatebox[origin=c]{90}{ETTm2}} & 96  &       0.189 &  0.282 &  \textbf{0.187} &  \textbf{0.274} &      \textbf{1.058} &      \textbf{2.837} \\
        & 192 &       0.258 &  0.326 &  \textbf{0.255} &  \textbf{0.314} &      \textbf{1.163} &      \textbf{3.681} \\
        & 336 &       0.323 &  0.363 &  \textbf{0.311} &  \textbf{0.350} &      \textbf{3.715} &      \textbf{3.581} \\
        & 720 &       0.425 &  0.421 &  \textbf{0.406} &  \textbf{0.403} &      \textbf{4.471} &      \textbf{4.276} \\ \cmidrule(lr){1-8}
    \multirow{4}{*}{\rotatebox[origin=c]{90}{Electricity}} & 96  &       \textbf{0.185} &  \textbf{0.300} &  \textbf{0.185} &  \textbf{0.300} &      \textbf{0.000} &      \textbf{0.000} \\
        & 192 &       \textbf{0.201} &  \textbf{0.316} &  \textbf{0.201} &  \textbf{0.316} &      \textbf{0.000} &      \textbf{0.000} \\
\bottomrule
\end{tabular}}

\end{minipage}
\begin{minipage}[c]{0.5\textwidth}
\centering
\caption{FEDformer-w multivariate}
\label{tab:fedformerw_mul}
    \scalebox{0.67}{
    \begin{tabular}{ll|rr|rrrr}
    \toprule
        & &  \multicolumn{2}{c|}{FEDformer-w} & \multicolumn{4}{c}{TSAA} \\
        & &           MSE &    MAE &    MSE$\downarrow$ &    MAE$\downarrow$ & MSE$\%$$\uparrow$  & MAE$\%$$\uparrow$  \\
    \midrule
    \multirow{4}{*}{\rotatebox[origin=c]{90}{ETTm2}} & 96  &       0.205 &  0.290 &  \textbf{0.199} &  \textbf{0.280} &      \textbf{2.927} &      \textbf{3.448} \\
        & 192 &       0.270 &  0.329 &  \textbf{0.255} &  \textbf{0.314} &      \textbf{5.556} &      \textbf{4.559} \\
        & 336 &       0.328 &  0.364 &  \textbf{0.316} &  \textbf{0.352} &      \textbf{3.659} &      \textbf{3.297} \\
        & 720 &       0.433 &  0.425 &  \textbf{0.410} &  \textbf{0.404} &      \textbf{5.312} &      \textbf{4.941} \\ \cmidrule(lr){1-8}
    \multirow{4}{*}{\rotatebox[origin=c]{90}{Electricity}} & 96  &       0.196 &  0.310 &  \textbf{0.183} &  \textbf{0.297} &      \textbf{6.633} &      \textbf{4.194} \\
        & 192 &       0.199 &  0.310 &  \textbf{0.195} &  \textbf{0.309} &      \textbf{2.010} &      \textbf{0.323} \\
\bottomrule
\end{tabular}}

\end{minipage}
\end{table*}

\begin{table*}[ht]

\begin{minipage}[c]{0.5\textwidth}
\centering
\caption{Informer univariate}
\label{tab:informer_uni}
\scalebox{0.67}{

\begin{tabular}{ll|rr|rrrr}
    \toprule
        & &  \multicolumn{2}{c|}{Informer} & \multicolumn{4}{c}{TSAA} \\
        & &       MSE &    MAE &    MSE$\downarrow$ &    MAE$\downarrow$ & MSE$\%$$\uparrow$  & MAE$\%$$\uparrow$  \\
    \midrule
    \multirow{4}{*}{\rotatebox[origin=c]{90}{ETTm2}} & 96  &    \textbf{0.085} &  \textbf{0.225} &  \textbf{0.085} &  0.226 &      \textbf{0.000} &     -0.444 \\
            & 192 &    0.130 &  0.282 &  \textbf{0.128} &  \textbf{0.279} &      \textbf{1.538} &     \textbf{ 1.064} \\
            & 336 &    \textbf{0.161} &  \textbf{0.314} &  0.164 &  0.317 &     -1.863 &     -0.955 \\
            & 720 &    0.221 &  0.373 &  \textbf{0.221} &  \textbf{0.371} &      \textbf{0.000} &      \textbf{0.536} \\ \cmidrule(lr){1-8}
    \multirow{4}{*}{\rotatebox[origin=c]{90}{Electricity}} & 96  &    0.261 &  0.367 &  \textbf{0.245} &  \textbf{0.356} &      \textbf{6.130} &      \textbf{2.997} \\
            & 192 &    \textbf{0.285} &  \textbf{0.386} &  0.297 &  0.401 &     -4.211 &     -3.886 \\
\bottomrule
\end{tabular}}

\end{minipage}
\begin{minipage}[c]{0.5\textwidth}
\centering
\caption{Autoformer univariate}
\label{tab:autoformer_uni}
    \scalebox{0.67}{
    \begin{tabular}{ll|rr|rrrr}
    \toprule
        & &  \multicolumn{2}{c|}{Autoformer} & \multicolumn{4}{c}{TSAA} \\
        & &           MSE &    MAE &    MSE$\downarrow$ &    MAE$\downarrow$ & MSE$\%$$\uparrow$  & MAE$\%$$\uparrow$  \\
    \midrule
    \multirow{4}{*}{\rotatebox[origin=c]{90}{ETTm2}} & 96  &      0.123 &  0.270 &  \textbf{0.106} &  \textbf{0.250} &     \textbf{13.821} &      \textbf{7.407} \\
            & 192 &      0.141 &  0.289 &  \textbf{0.126} &  \textbf{0.273} &     \textbf{10.638} &      \textbf{5.536} \\
            & 336 &      0.170 &  0.319 &  \textbf{0.139} &  \textbf{0.290} &     \textbf{18.235} &      \textbf{9.091} \\
            & 720 &      0.206 &  0.353 &  \textbf{0.187} &  \textbf{0.338} &      \textbf{9.223} &      \textbf{4.249} \\ \cmidrule(lr){1-8}
    \multirow{4}{*}{\rotatebox[origin=c]{90}{Electricity}} & 96  &      0.454 &  0.508 &  \textbf{0.388} &  \textbf{0.456} &     \textbf{14.537} &     \textbf{10.236} \\
            & 192 &      0.511 &  0.532 &  \textbf{0.463} &  \textbf{0.509} &      \textbf{9.393} &      \textbf{4.323} \\
\bottomrule
\end{tabular}}

\end{minipage}

\begin{minipage}[c]{0.5\textwidth}
\centering
\caption{FEDformer-f univariate}
\label{tab:fedformerf_uni}
\scalebox{0.67}{

\begin{tabular}{ll|rr|rrrr}
    \toprule
        & &  \multicolumn{2}{c|}{FEDformer-f} & \multicolumn{4}{c}{TSAA} \\
        & &       MSE &    MAE &    MSE$\downarrow$ &    MAE$\downarrow$ & MSE$\%$$\uparrow$  & MAE$\%$$\uparrow$  \\
    \midrule
    \multirow{4}{*}{\rotatebox[origin=c]{90}{ETTm2}} & 96  &       \textbf{0.068} &  \textbf{0.198} &  \textbf{0.068} &  \textbf{0.198} &      \textbf{0.000} &      \textbf{0.000} \\
            & 192 &       \textbf{0.096} &  \textbf{0.238} &  \textbf{0.096} &  \textbf{0.238} &      \textbf{0.000} &      \textbf{0.000} \\
            & 336 &       \textbf{0.138} &  \textbf{0.286} &  0.140 &  0.290 &     -1.449 &     -1.399 \\
            & 720 &       \textbf{0.189} &  \textbf{0.335} &  0.190 &  0.336 &     -0.529 &     -0.299 \\ \cmidrule(lr){1-8}
    \multirow{4}{*}{\rotatebox[origin=c]{90}{Electricity}} & 96  &       0.244 &  0.364 &  \textbf{0.244} &  \textbf{0.356} &      \textbf{0.000} &      \textbf{2.198} \\
            & 192 &       \textbf{0.276} &  0.382 &  0.277 &  \textbf{0.381} &     -0.362 &      \textbf{0.262} \\
\bottomrule
\end{tabular}}

\end{minipage}
\begin{minipage}[c]{0.5\textwidth}
\centering
\caption{NBEATS-I univariate}
\label{tab:nbeatsi_uni}
    \scalebox{0.67}{
    \begin{tabular}{ll|rr|rrrr}
    \toprule
        & &  \multicolumn{2}{c|}{NBEATS-I} & \multicolumn{4}{c}{TSAA} \\
        & &           MSE &    MAE &    MSE$\downarrow$ &    MAE$\downarrow$ & MSE$\%$$\uparrow$  & MAE$\%$$\uparrow$  \\
    \midrule
        \multirow{4}{*}{\rotatebox[origin=c]{90}{ETTm2}} & 96  &    0.080 &  0.213 &  \textbf{0.075} &  \textbf{0.202} &      \textbf{6.250} &     \textbf{ 5.164} \\
            & 192 &    0.103 &  0.240 &  \textbf{0.102} &  \textbf{0.237} &      \textbf{0.971} &      \textbf{1.250} \\
            & 336 &    \textbf{0.162} &  \textbf{0.312} &  0.171 &  0.321 &     -5.556 &     -2.885 \\
            & 720 &    \textbf{0.199} &  \textbf{0.347} &  0.231 &  0.376 &    -16.080 &     -8.357 \\ \cmidrule(lr){1-8}
    \multirow{4}{*}{\rotatebox[origin=c]{90}{Electricity}}  & 96  &    0.326 &  0.402 &  \textbf{0.270} &  \textbf{0.360} &     \textbf{17.178} &     \textbf{10.448} \\
            & 192 &    0.350 &  0.417 &  \textbf{0.277} &  \textbf{0.368} &     \textbf{20.857} &     \textbf{11.751} \\
\bottomrule
\end{tabular}}

\end{minipage}

\begin{minipage}[c]{0.5\textwidth}
\centering
\caption{NBEATS-G univariate}
\label{tab:nbeatsg_uni}
\scalebox{0.670}{

\begin{tabular}{ll|rr|rrrr}
    \toprule
        & &  \multicolumn{2}{c|}{NBEATS-G} & \multicolumn{4}{c}{TSAA} \\
        & &       MSE &    MAE &    MSE$\downarrow$ &    MAE$\downarrow$ & MSE$\%$$\uparrow$  & MAE$\%$$\uparrow$  \\
    \midrule
            \multirow{4}{*}{\rotatebox[origin=c]{90}{ETTm2}} & 96  &    0.080 &  0.210 &  \textbf{0.071} &  \textbf{0.192} &     \textbf{11.250} &      \textbf{8.571} \\
            & 192 &    0.110 &  0.250 &  \textbf{0.109} &  \textbf{0.246} &      \textbf{0.909} &      \textbf{1.600} \\
            & 336 &    \textbf{0.172} &  \textbf{0.320} &  0.176 &  0.323 &     -2.326 &     -0.938 \\
            & 720 &    \textbf{0.201} &  \textbf{0.353} &  0.218 &  0.366 &     -8.458 &     -3.683 \\ \cmidrule(lr){1-8}
            \multirow{4}{*}{\rotatebox[origin=c]{90}{Electricity}} & 96  &    0.324 &  0.397 &  \textbf{0.263} &  \textbf{0.354} &   \textbf{18.827} &   \textbf{10.831} \\
            & 192 &    0.363 &  0.420 &  \textbf{0.278} &  \textbf{0.368} &    \textbf{ 23.416} &     \textbf{12.381} \\
\bottomrule
\end{tabular}}

\end{minipage}
\begin{minipage}[c]{0.5\textwidth}
\centering

\end{minipage}
\end{table*}

\clearpage

\subsection{Main results with standard deviation}
\label{app:std_results}

The following tables Tabs.~\ref{tab:multi_results_std1}, \ref{tab:uni_results_std1} augment the tables presented in Sec.~\ref{sec:results} in the main text with standard deviation measures computed over three different seed numbers.

\begin{table*}[h!]
    \caption{Multivariate long-term time-series forecasting results with the standard deviation.}
    \label{tab:multi_results_std1}
\scalebox{0.8}{
\begin{tabular}{ll|rr|rr|rr|rr|rr}
    \toprule
        & & \multicolumn{2}{c|}{Informer} & \multicolumn{2}{c|}{Autoformer} & \multicolumn{2}{c|}{FEDformer-w} & \multicolumn{2}{c|}{FEDformer-f}  \\
        & &      \multicolumn{1}{c}{MSE} & \multicolumn{1}{c|}{MAE} &  \multicolumn{1}{c}{MSE}  & \multicolumn{1}{c|}{MAE} &        \multicolumn{1}{c}{MSE}  & \multicolumn{1}{c|}{MAE} &         \multicolumn{1}{c}{MSE}  & \multicolumn{1}{c|}{MAE} \\
    \midrule
\multirow{4}{*}{\rotatebox[origin=c]{90}{ETTm2}} & 96  &  0.545 $\pm$ 0.024 &  0.588 $\pm$ 0.014 &  0.231 $\pm$ 0.004 &     0.31 $\pm$ 0.0 &  0.205 $\pm$ 0.001 &     0.29 $\pm$ 0.0 &  0.189 $\pm$ 0.001 &  0.282 $\pm$ 0.001 \\
    & 192 &  1.054 $\pm$ 0.044 &  0.808 $\pm$ 0.018 &  0.289 $\pm$ 0.013 &   0.346 $\pm$ 0.01 &     0.27 $\pm$ 0.0 &    0.329 $\pm$ 0.0 &  0.258 $\pm$ 0.002 &  0.326 $\pm$ 0.001 \\
    & 336 &  1.523 $\pm$ 0.036 &  0.948 $\pm$ 0.014 &  0.341 $\pm$ 0.008 &  0.375 $\pm$ 0.004 &    0.328 $\pm$ 0.0 &    0.364 $\pm$ 0.0 &  0.323 $\pm$ 0.004 &  0.363 $\pm$ 0.004 \\
    & 720 &  3.878 $\pm$ 0.072 &  1.474 $\pm$ 0.004 &  0.444 $\pm$ 0.005 &  0.434 $\pm$ 0.005 &  0.433 $\pm$ 0.005 &  0.425 $\pm$ 0.005 &  0.425 $\pm$ 0.008 &  0.421 $\pm$ 0.001  \\
    \cmidrule(lr){1-10}
\multirow{4}{*}{\rotatebox[origin=c]{90}{Electricity}} & 96  &  0.336 $\pm$ 0.013 &  0.416 $\pm$ 0.007 &    0.2 $\pm$ 0.007 &  0.316 $\pm$ 0.007 &  0.196 $\pm$ 0.002 &   0.31 $\pm$ 0.002 &  0.185 $\pm$ 0.001 &    0.3 $\pm$ 0.001  \\
    & 192 &   0.36 $\pm$ 0.015 &  0.441 $\pm$ 0.012 &    0.217 $\pm$ 0.0 &  0.326 $\pm$ 0.002 &  0.199 $\pm$ 0.001 &   0.31 $\pm$ 0.002 &  0.201 $\pm$ 0.007 &  0.316 $\pm$ 0.008    \\
\bottomrule
    \end{tabular}}

    \label{tab:multi_results_std2}
\scalebox{0.8}{
    \begin{tabular}{ll|rr|rr|rr|rr|rr|rr}
    \toprule
        & &  \multicolumn{2}{c|}{TSAA} \\
        & &   \multicolumn{1}{c}{MSE}  &  \multicolumn{1}{c|}{MAE}  \\
    \midrule
\multirow{4}{*}{\rotatebox[origin=c]{90}{ETTm2}} & 96  &    0.187 $\pm$ 0.000 &    0.274 $\pm$ 0.000 \\
    & 192  &  0.255 $\pm$ 0.001 &  0.314 $\pm$ 0.001 \\
    & 336 &  0.304 $\pm$ 0.009 &     0.350 $\pm$ 0.002 \\
    & 720  &  0.398 $\pm$ 0.007 &  0.403 $\pm$ 0.002 \\
    \cmidrule(lr){1-4}
\multirow{4}{*}{\rotatebox[origin=c]{90}{Electricity}} & 96  &  0.183 $\pm$ 0.001 &  0.297 $\pm$ 0.001 \\
    & 192 &  0.195 $\pm$ 0.001 &  0.309 $\pm$ 0.002 \\
\bottomrule
    \end{tabular}}
\end{table*} 
\clearpage

\begin{table*}[ht]
    \caption{Univariate long-term time-series forecasting results with the standard deviation.}
    \label{tab:uni_results_std1}
\scalebox{0.8}{
    \begin{tabular}{ll|rr|rr|rr|rr|rr|rr|rr}
    \toprule
        & & \multicolumn{2}{c|}{Informer} & \multicolumn{2}{c|}{Autoformer} & \multicolumn{2}{c|}{FEDformer-f}& \multicolumn{2}{c|}{N-BEATS-I} \\
        &  &  \multicolumn{1}{c}{MSE}  & \multicolumn{1}{c|}{MAE} &   \multicolumn{1}{c}{MSE}  & \multicolumn{1}{c|}{MAE} &   \multicolumn{1}{c}{MSE}  & \multicolumn{1}{c|}{MAE} &  \multicolumn{1}{c}{MSE}  & \multicolumn{1}{c|}{MAE}  \\
\midrule
\multirow{4}{*}{\rotatebox[origin=c]{90}{ETTm2}} & 96  &  0.085 $\pm$ 0.004 &  0.225 $\pm$ 0.006 &  0.123 $\pm$ 0.017 &   0.27 $\pm$ 0.018 &  0.068 $\pm$ 0.001 &  0.198 $\pm$ 0.002 &   0.08 $\pm$ 0.003 &  0.213 $\pm$ 0.006  \\
        & 192 &   0.13 $\pm$ 0.007 &  0.282 $\pm$ 0.008 &   0.141 $\pm$ 0.01 &   0.289 $\pm$ 0.01 &  0.096 $\pm$ 0.001 &  0.238 $\pm$ 0.001 &  0.103 $\pm$ 0.004 &   0.24 $\pm$ 0.006  \\
        & 336 &  0.161 $\pm$ 0.008 &  0.314 $\pm$ 0.006 &   0.17 $\pm$ 0.046 &  0.319 $\pm$ 0.042 &  0.138 $\pm$ 0.013 &  0.286 $\pm$ 0.014 &  0.162 $\pm$ 0.009 &  0.312 $\pm$ 0.009  \\
        & 720 &  0.221 $\pm$ 0.006 &  0.373 $\pm$ 0.007 &   0.206 $\pm$ 0.02 &  0.353 $\pm$ 0.017 &  0.189 $\pm$ 0.002 &  0.335 $\pm$ 0.002 &  0.199 $\pm$ 0.007 &  0.347 $\pm$ 0.007 \\ \cmidrule(lr){1-10}
\multirow{4}{*}{\rotatebox[origin=c]{90}{Electricity}} & 96  &  0.261 $\pm$ 0.005 &  0.367 $\pm$ 0.002 &  0.454 $\pm$ 0.014 &  0.508 $\pm$ 0.014 &  0.244 $\pm$ 0.001 &  0.364 $\pm$ 0.002 &  0.326 $\pm$ 0.006 &  0.402 $\pm$ 0.004  \\
        & 192 &  0.285 $\pm$ 0.006 &  0.386 $\pm$ 0.003 &   0.511 $\pm$ 0.05 &  0.532 $\pm$ 0.027 &  0.276 $\pm$ 0.004 &  0.382 $\pm$ 0.004 &   0.35 $\pm$ 0.008 &  0.417 $\pm$ 0.005  \\
\bottomrule
    \end{tabular}}

    \label{tab:uni_results_std2}
\scalebox{0.8}{
    \begin{tabular}{ll|rr|rr|rr|}
    \toprule
        & & \multicolumn{2}{c|}{N-BEATS-G} &  \multicolumn{2}{c|}{TSAA} \\
        &  &  \multicolumn{1}{c}{MSE}   & \multicolumn{1}{c|}{MAE} & \multicolumn{1}{c}{MSE}  & \multicolumn{1}{c|}{MAE} \\
\midrule
\multirow{4}{*}{\rotatebox[origin=c]{90}{ETTm2}} & 96  &   0.08 $\pm$ 0.005 &   0.21 $\pm$ 0.007  &  0.068 $\pm$ 0.001 &  0.192 $\pm$ 0.002 \\
        & 192 &   0.11 $\pm$ 0.004 &   0.25 $\pm$ 0.005  &  0.096 $\pm$ 0.001 &  0.237 $\pm$ 0.004 \\
        & 336 &    0.172 $\pm$ 0.007 &   0.32 $\pm$ 0.007  &  0.139 $\pm$ 0.005 &   0.29 $\pm$ 0.005 \\
        & 720 &   0.201 $\pm$ 0.008 &  0.353 $\pm$ 0.008  &    0.187 $\pm$ 0.008 &  0.336 $\pm$ 0.001 \\ \cmidrule(lr){1-6}
\multirow{4}{*}{\rotatebox[origin=c]{90}{Electricity}} & 96  &  0.324 $\pm$ 0.005 &  0.397 $\pm$ 0.002  &  0.244 $\pm$ 0.006 &  0.354 $\pm$ 0.012 \\
        & 192 &    0.363 $\pm$ 0.005 &   0.42 $\pm$ 0.003  &  0.277 $\pm$ 0.003 &  0.368 $\pm$ 0.003 \\
\bottomrule
    \end{tabular}}
\end{table*}

\subsection{Comparing with Crossformer, PatchTST, and iTransformer models}

Multivariate results for TSAA using the baselines: Crossformer, PatchTST, and iTransformer. Each result reports an average of three random seeds, with a lookback of 96. The results are presented in Tabs.~\ref{tab:cross_mul}, \ref{tab:patchtst_mul}, and \ref{tab:itransformer_mul}.

\begin{table*}[h!]
\begin{minipage}[c]{0.5\textwidth}
    \centering
    \caption{Crossformer multivariate}
    \label{tab:cross_mul}
    \scalebox{0.67}{
    \begin{tabular}{ll|rr|rrrr}
     \toprule 
     & &  \multicolumn{2}{c|}{Crossformer} & \multicolumn{4}{c}{TSAA} \\
        &     &  MSE &  MAE &  MSE$\downarrow$ &  MAE$\downarrow$ &  MSE$\%$$\uparrow$   &  MAE$\%$$\uparrow$   \\
    \midrule
     \cmidrule(lr){1-8} 
     \multirow{4}{*}{\rotatebox[origin=c]{90}{ETTm2}} & 96  &           0.263 &            0.35 &  \textbf{0.197} &  \textbf{0.283} &  \textbf{25.095} &  \textbf{19.143} \\
        & 192 &           0.476 &           0.492 &  \textbf{0.273} &  \textbf{0.336} &  \textbf{42.647} &  \textbf{31.707} \\
        & 336 &           0.855 &           0.638 &  \textbf{0.335} &  \textbf{0.378} &  \textbf{60.819} &  \textbf{40.752} \\
        & 720 &           3.293 &            1.26 &  \textbf{0.397} &    \textbf{0.4} &  \textbf{87.944} &  \textbf{68.254} \\
     \cmidrule(lr){1-8} 
     \multirow{4}{*}{\rotatebox[origin=c]{90}{ECL}} & 96  &           0.144 &           0.243 &  \textbf{0.141} &   \textbf{0.24} &   \textbf{2.083} &   \textbf{1.235} \\
        & 192 &           0.165 &           0.264 &  \textbf{0.157} &  \textbf{0.254} &   \textbf{4.848} &   \textbf{3.788} \\
\bottomrule
    \end{tabular}}
\end{minipage}
\begin{minipage}[c]{0.5\textwidth}
\centering
\caption{PatchTST multivariate}
\label{tab:patchtst_mul}
    \scalebox{0.67}{
\begin{tabular}{ll|rr|rrrr}
 \toprule 
 & &  \multicolumn{2}{c|}{PatchTST} & \multicolumn{4}{c}{TSAA} \\
    &     &  MSE &  MAE &  MSE$\downarrow$ &  MAE$\downarrow$ &  MSE$\%$$\uparrow$   &  MAE$\%$$\uparrow$   \\
\midrule
 \cmidrule(lr){1-8} 
 \multirow{4}{*}{\rotatebox[origin=c]{90}{ETTm2}} & 96  &  \textbf{0.175} &           0.259 &  \textbf{0.175} &  \textbf{0.256} &              0.0 &  \textbf{1.158} \\
    & 192 &           0.242 &           0.303 &   \textbf{0.24} &  \textbf{0.301} &   \textbf{0.826} &   \textbf{0.66} \\
    & 336 &           0.303 &           0.341 &  \textbf{0.302} &  \textbf{0.339} &    \textbf{0.33} &  \textbf{0.587} \\
    & 720 &           0.401 &           0.399 &    \textbf{0.4} &  \textbf{0.396} &   \textbf{0.249} &  \textbf{0.752} \\
 \cmidrule(lr){1-8} 
 \multirow{4}{*}{\rotatebox[origin=c]{90}{ECL}} & 96  &  \textbf{0.187} &  \textbf{0.269} &  \textbf{0.187} &  \textbf{0.269} &              0.0 &             0.0 \\
    & 192 &  \textbf{0.191} &  \textbf{0.274} &  \textbf{0.191} &  \textbf{0.274} &              0.0 &             0.0 \\
\bottomrule
\end{tabular}}
\end{minipage}
\begin{minipage}[c]{0.5\textwidth}
\centering
\caption{iTransformer multivariate}
\label{tab:itransformer_mul}
\scalebox{0.670}{

\begin{tabular}{ll|rr|rrrr}
 \toprule 
 & &  \multicolumn{2}{c|}{iTransformer} & \multicolumn{4}{c}{TSAA} \\
    &     &  MSE &  MAE &  MSE$\downarrow$ &  MAE$\downarrow$ &  MSE$\%$$\uparrow$   &  MAE$\%$$\uparrow$   \\
\midrule
 \cmidrule(lr){1-8} 
 \multirow{4}{*}{\rotatebox[origin=c]{90}{ETTm2}} & 96  &           0.184 &            0.27 &   \textbf{0.18} &  \textbf{0.259} &   \textbf{2.174} &  \textbf{4.074} \\
    & 192 &           0.251 &           0.312 &  \textbf{0.246} &  \textbf{0.302} &   \textbf{1.992} &  \textbf{3.205} \\
    & 336 &           0.315 &           0.351 &  \textbf{0.311} &  \textbf{0.345} &    \textbf{1.27} &  \textbf{1.709} \\
    & 720 &            0.41 &           0.405 &  \textbf{0.403} &  \textbf{0.397} &   \textbf{1.707} &  \textbf{1.975} \\
 \cmidrule(lr){1-8} 
 \multirow{4}{*}{\rotatebox[origin=c]{90}{ECL}} & 96  &  \textbf{0.147} &  \textbf{0.239} &  \textbf{0.147} &  \textbf{0.239} &              0.0 &             0.0 \\
    & 192 &  \textbf{0.163} &  \textbf{0.255} &  \textbf{0.163} &  \textbf{0.255} &              0.0 &             0.0 \\
\bottomrule
\end{tabular}}

\end{minipage}
\begin{minipage}[c]{0.5\textwidth}
\centering
\end{minipage}
\end{table*}

\subsection{Autoformer and FEDformer augmented with the best augmentations}

We show in Tabs.~\ref{tab:best_da_auto} and \ref{tab:best_da_fed} a comparison between TSAA and Autoformer and FEDformer, respectively. Specifically, we train from scratch Autoformer and FEDformer only using the best augmentations found by TSAA. Overall, the results are inconsistent, highlighting that in some cases, these augmentations yield strong results, but in other cases, results are inferior. In comparison, TSAA is consistent across all models and datasets.

\begin{table*}[t]
\centering
    \caption{ Autoformer: TSAA compared to the direct application of the best-performing transformations trend downscaling (TD), jittering, mixup, and smoothing. Each transformation was deployed with a fixed magnitude of 0.5.}
    \label{tab:best_da_auto}
   \scalebox{0.65}{
\begin{tabular}{ll|cc|cc|cc|cc|cc}
\toprule
    &      &     TD MSE &     TD MAE & Jittering  MSE & Jittering MAE &  Mixup MSE &  Mixup MAE & Smooth MSE & Smooth MAE &           TSAA MSE &           TSAA MAE \\
{} & pred\_len &                    &                    &                       &                       &                    &                    &                    &                    &                    &                    \\
\midrule
 \cmidrule(lr){1-12} 
 \multirow{4}{*}{\rotatebox[origin=c]{90}{ETTm2}} & 96 &              0.253 &              0.324 &                 0.265 &                  0.33 &  \underline{0.236} &  \underline{0.317} &              0.278 &              0.337 &     \textbf{0.211} &     \textbf{0.293} \\
    & 192 &              0.344 &              0.368 &                  0.29 &                 0.344 &  \underline{0.277} &  \underline{0.335} &              0.293 &              0.349 &     \textbf{0.269} &     \textbf{0.327} \\
    & 336 &              0.353 &              0.384 &                 0.419 &                 0.418 &              0.341 &              0.371 &  \underline{0.334} &   \underline{0.37} &     \textbf{0.322} &     \textbf{0.359} \\
    & 720 &              0.439 &              0.432 &                 0.473 &                 0.452 &  \underline{0.426} &  \underline{0.417} &              0.433 &              0.426 &      \textbf{0.41} &     \textbf{0.407} \\
    \hline 
 & Avg. &              0.347 &              0.377 &                 0.362 &                 0.386 &   \underline{0.32} &   \underline{0.36} &              0.334 &               0.37 &     \textbf{0.303} &     \textbf{0.346} \\
 \hline \hline 
 \multirow{4}{*}{\rotatebox[origin=c]{90}{Traffic}} & 96 &              0.632 &              0.402 &     \underline{0.623} &     \underline{0.385} &              0.633 &              0.401 &              0.634 &              0.396 &     \textbf{0.602} &     \textbf{0.375} \\
    & 192 &  \underline{0.659} &  \underline{0.415} &                 0.667 &                 0.416 &              0.742 &              0.463 &     \textbf{0.632} &     \textbf{0.401} &              0.663 &              0.416 \\
    & 336 &              0.645 &              0.402 &     \underline{0.634} &     \underline{0.398} &              0.683 &              0.429 &              0.654 &              0.408 &     \textbf{0.627} &     \textbf{0.387} \\
    & 720 &              0.679 &              0.416 &                 0.665 &                 0.407 &              0.744 &              0.455 &     \textbf{0.661} &  \underline{0.406} &  \underline{0.662} &     \textbf{0.405} \\
    \hline 
 & Avg. &              0.654 &              0.409 &                 0.647 &     \underline{0.401} &                0.7 &              0.437 &  \underline{0.645} &              0.403 &     \textbf{0.639} &     \textbf{0.396} \\
 \hline \hline 
 \multirow{4}{*}{\rotatebox[origin=c]{90}{Weather}} & 96 &              0.242 &              0.317 &                 0.259 &                 0.328 &  \underline{0.222} &  \underline{0.297} &              0.246 &               0.32 &     \textbf{0.216} &     \textbf{0.292} \\
    & 192 &              0.295 &              0.355 &                 0.324 &                 0.375 &  \underline{0.286} &  \underline{0.346} &              0.331 &              0.382 &     \textbf{0.278} &     \textbf{0.336} \\
    & 336 &              0.367 &                0.4 &                 0.358 &                 0.391 &     \textbf{0.327} &     \textbf{0.367} &              0.379 &              0.408 &  \underline{0.341} &  \underline{0.381} \\
    & 720 &              0.436 &              0.443 &                 0.428 &                  0.43 &     \textbf{0.396} &  \underline{0.411} &              0.449 &              0.449 &  \underline{0.397} &      \textbf{0.41} \\
    \hline 
 & Avg. &  \underline{0.335} &  \underline{0.379} &                 0.342 &                 0.381 &     \textbf{0.308} &     \textbf{0.355} &              0.351 &               0.39 &     \textbf{0.308} &     \textbf{0.355} \\
 \hline \hline 
 \multirow{4}{*}{\rotatebox[origin=c]{90}{ILI}} & 24 &     \textbf{3.398} &     \textbf{1.284} &                 3.656 &                 1.311 &              3.709 &               1.35 &  \underline{3.463} &  \underline{1.289} &              3.565 &              1.302 \\
    & 36 &              2.897 &              1.117 &                 2.878 &                 1.096 &              3.024 &              1.135 &  \underline{2.812} &  \underline{1.087} &     \textbf{2.754} &     \textbf{1.068} \\
    & 48 &              2.879 &              1.126 &                  2.94 &                 1.132 &              3.048 &              1.171 &     \textbf{2.841} &     \textbf{1.111} &  \underline{2.856} &  \underline{1.114} \\
    & 60 &               2.85 &              1.135 &                 2.849 &     \underline{1.118} &              3.106 &              1.198 &     \textbf{2.805} &     \textbf{1.114} &  \underline{2.826} &  \underline{1.118} \\
    \hline 
 & Avg. &              3.006 &              1.166 &                 3.081 &     \underline{1.164} &              3.222 &              1.214 &      \textbf{2.98} &      \textbf{1.15} &    \underline{3.0} &      \textbf{1.15} \\
\bottomrule
\end{tabular}}
\end{table*}

\begin{table*}[t]
\centering
    \caption{FEDformer: TSAA compared to the direct application of the best-performing transformations trend downscaling (TD), jittering, mixup, and smoothing. Each transformation was deployed with a fixed magnitude of 0.5.}
    \label{tab:best_da_fed}
   \scalebox{0.65}{
\begin{tabular}{ll|cc|cc|cc|cc|cc}
\toprule
    &      &     TD MSE &     TD MAE & Jittering  MSE & Jittering MAE &  Mixup MSE &  Mixup MAE & Smooth MSE & Smooth MAE &           TSAA MSE &           TSAA MAE \\
{} & pred\_len &                    &                    &                       &                       &                    &                    &                    &                    &                    &                    \\
\midrule
 \cmidrule(lr){1-12} 
 \multirow{4}{*}{\rotatebox[origin=c]{90}{ETTm2}} & 96 &  \underline{0.187} &     \textbf{0.274} &     \underline{0.187} &                 0.279 &     \textbf{0.185} &  \underline{0.276} &              0.192 &              0.285 &  \underline{0.187} &     \textbf{0.274} \\
    & 192 &  \underline{0.253} &  \underline{0.315} &                 0.254 &                 0.321 &     \textbf{0.252} &              0.318 &              0.264 &               0.33 &              0.255 &     \textbf{0.314} \\
    & 336 &  \underline{0.313} &  \underline{0.354} &                 0.322 &                 0.366 &               0.32 &              0.358 &              0.324 &              0.365 &     \textbf{0.311} &      \textbf{0.35} \\
    & 720 &  \underline{0.409} &  \underline{0.409} &                 0.427 &                 0.425 &               0.42 &              0.414 &              0.424 &               0.42 &     \textbf{0.406} &     \textbf{0.403} \\
    \hline 
 & Avg. &      \textbf{0.29} &  \underline{0.338} &                 0.298 &                 0.348 &  \underline{0.294} &              0.342 &              0.301 &               0.35 &      \textbf{0.29} &     \textbf{0.335} \\
 \hline \hline 
 \multirow{4}{*}{\rotatebox[origin=c]{90}{Traffic}} & 96 &              0.596 &              0.382 &        \textbf{0.575} &         \textbf{0.36} &              0.592 &              0.379 &              0.592 &              0.368 &  \underline{0.577} &  \underline{0.362} \\
    & 192 &              0.629 &              0.397 &     \underline{0.604} &     \underline{0.373} &              0.619 &              0.393 &              0.612 &              0.378 &     \textbf{0.601} &     \textbf{0.371} \\
    & 336 &              0.635 &              0.401 &      \underline{0.62} &     \underline{0.384} &              0.635 &              0.403 &              0.629 &              0.386 &     \textbf{0.619} &     \textbf{0.383} \\
    & 720 &              0.646 &                0.4 &        \textbf{0.629} &        \textbf{0.385} &              0.664 &              0.415 &              0.639 &  \underline{0.387} &  \underline{0.632} &              0.388 \\
    \hline 
 & Avg. &              0.626 &              0.395 &        \textbf{0.607} &        \textbf{0.376} &              0.628 &              0.398 &  \underline{0.618} &   \underline{0.38} &     \textbf{0.607} &     \textbf{0.376} \\
 \hline \hline 
 \multirow{4}{*}{\rotatebox[origin=c]{90}{Weather}} & 96 &              0.228 &              0.307 &                 0.283 &                 0.358 &     \textbf{0.199} &     \textbf{0.271} &              0.217 &              0.296 &  \underline{0.207} &  \underline{0.285} \\
    & 192 &              0.264 &              0.323 &                 0.314 &                 0.364 &     \textbf{0.249} &     \textbf{0.308} &              0.324 &              0.372 &  \underline{0.252} &  \underline{0.311} \\
    & 336 &              0.318 &              0.357 &                 0.368 &                 0.398 &     \textbf{0.305} &     \textbf{0.345} &              0.359 &              0.394 &  \underline{0.313} &  \underline{0.355} \\
    & 720 &              0.393 &              0.401 &                 0.417 &                 0.421 &     \textbf{0.377} &     \textbf{0.385} &              0.419 &              0.422 &  \underline{0.382} &  \underline{0.395} \\
    \hline 
 & Avg. &              0.301 &              0.347 &                 0.346 &                 0.385 &     \textbf{0.282} &     \textbf{0.327} &               0.33 &              0.371 &  \underline{0.288} &  \underline{0.336} \\
 \hline \hline 
 \multirow{4}{*}{\rotatebox[origin=c]{90}{ILI}} & 24 &              3.306 &              1.272 &                 3.272 &                 1.254 &              3.544 &              1.329 &  \underline{3.258} &  \underline{1.253} &      \textbf{3.15} &     \textbf{1.219} \\
    & 36 &              2.682 &              1.082 &     \underline{2.638} &     \underline{1.063} &              2.877 &              1.127 &              2.653 &              1.069 &     \textbf{2.578} &     \textbf{1.049} \\
    & 48 &              2.656 &              1.088 &      \underline{2.61} &        \textbf{1.067} &              2.844 &              1.135 &              2.618 &              1.073 &     \textbf{2.609} &  \underline{1.069} \\
    & 60 &               2.92 &              1.178 &                  2.87 &                 1.158 &              3.072 &              1.213 &  \underline{2.832} &  \underline{1.148} &     \textbf{2.805} &      \textbf{1.14} \\
    \hline 
 & Avg. &              2.891 &              1.155 &                 2.848 &     \underline{1.136} &              3.084 &              1.201 &   \underline{2.84} &  \underline{1.136} &     \textbf{2.786} &     \textbf{1.119} \\
\bottomrule
\end{tabular}}
\end{table*}

\subsection{Limitations of TSAA}

TSAA shows less favorable results when applied to the Exchange dataset. While several factors may contribute to this underperformance, we would like to focus on what we believe is the key distinction of this dataset compared to others, supported by additional experiments. The Exchange dataset captures foreign exchange prices between major currency pairs, categorizing it as financial data similar to stock market data. It is widely accepted that financial data often exhibits random fluctuations, characteristic of a random walk (RW) behavior, as suggested in \citep{fama1965behavior}. We hypothesize that the random walk component is dominant in the Exchange dataset, which may explain why TSAA encounters difficulties when applied to datasets with a strong random walk influence.
    
\begin{figure}[t]
    \centering
    \includegraphics[width=0.87\textwidth]{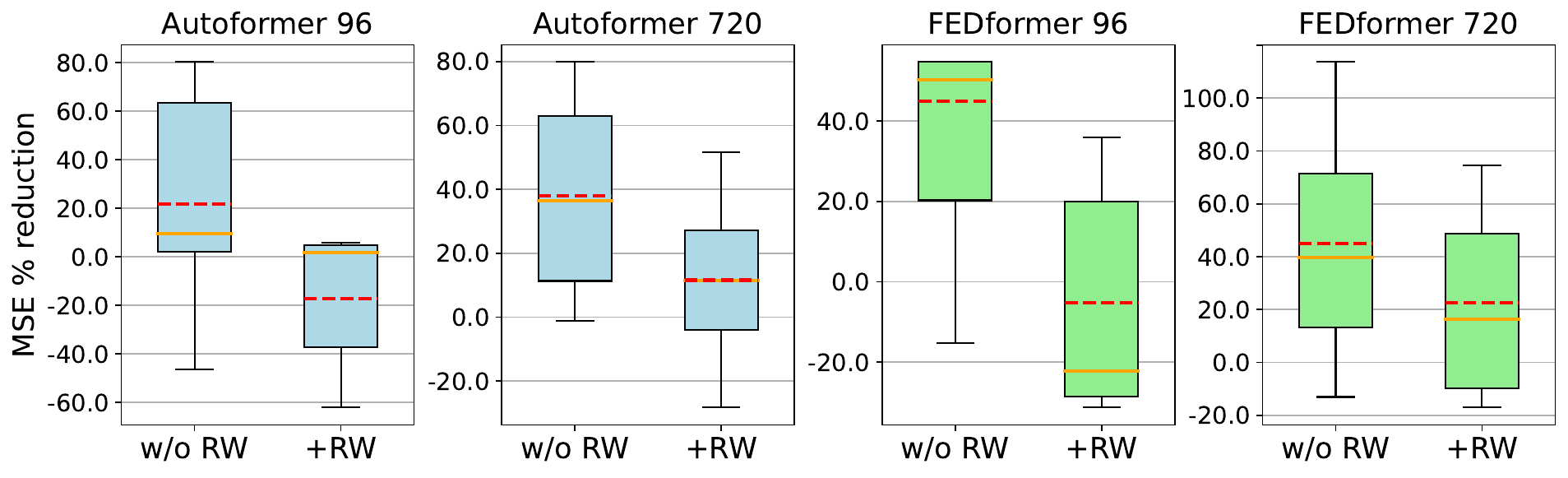} 
    \includegraphics[width=0.87\textwidth]{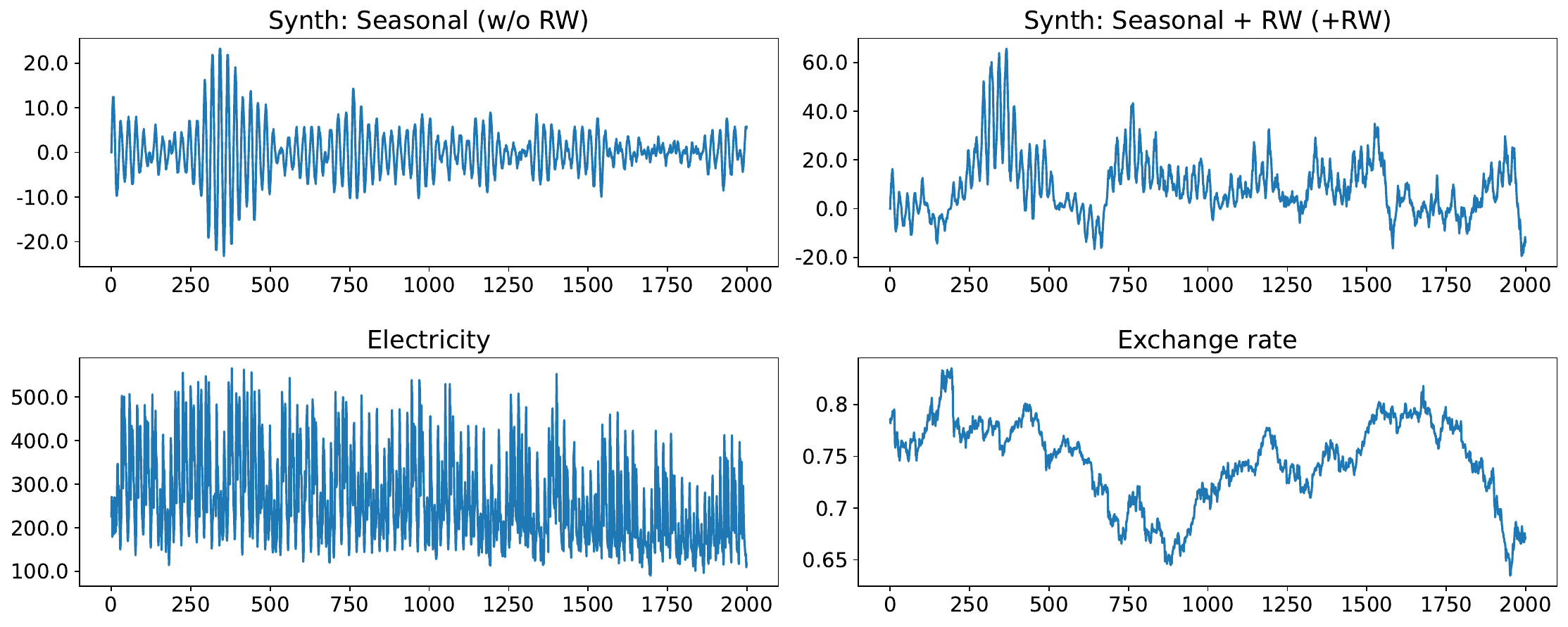} 
    \caption{Top: MSE \% reduction with TSAA for two synthetic datasets characterized by a seasonal component only (w/o RW), and a seasonal component with a random walk (+RW), respectively. The red dashed line and the orange line represent the mean and median respectively. The reported results represent the performance of five different setups. Bottom: Samples of the synthetic datasets used in the given experiments, and real datasets with similar corresponding characteristics.}
    \label{fig:limit_rw}
\end{figure}

To empirically validate our hypothesis, we evaluated the performance of TSAA on two different types of datasets: (1) A seasonal dataset (w/o RW), characterized by strong seasonal components without any random walk influence, and (2) A seasonal + random walk dataset (+RW), where a random walk component was added to the seasonal data. To further remove stationarity, both series were multiplied by an additional random walk vector, formally represented as: (1) $x_s * x_{rw}$, and (2) $(x_s + x_{rw}) * \hat{x}_{rw}$, where $x_s$ is the seasonal component, and $x_{rw}$ and $\hat{x}_{rw}$ are two independent random walk vectors, with the operation $*$ applied element-wise. Examples of these datasets are illustrated in Fig. \ref{fig:limit_rw}. The results indicate that, for both Autoformer and FEDformer, performance on the dataset containing a random walk component (+RW) is notably lower compared to the seasonal dataset (w/o RW), despite both sharing the same $x_s$ and $x_{rw}$ components. Specifically, the median and mean performance on the (w/o RW) dataset consistently surpass those of the (+RW) dataset across all configurations, as shown in Fig. \ref{fig:limit_rw}. In our experiment, five different setups using generated datasets were tested, each maintaining the same $x_s$, $x_{rw}$, and $\hat{x}_{rw}$ components.

\clearpage
\subsection{Further results}
In this section we provide bar comparisons of the results presented in \ref{app:ex_results}.

\begin{figure}[h]
    \centering
    \includegraphics[width=1.0\textwidth]{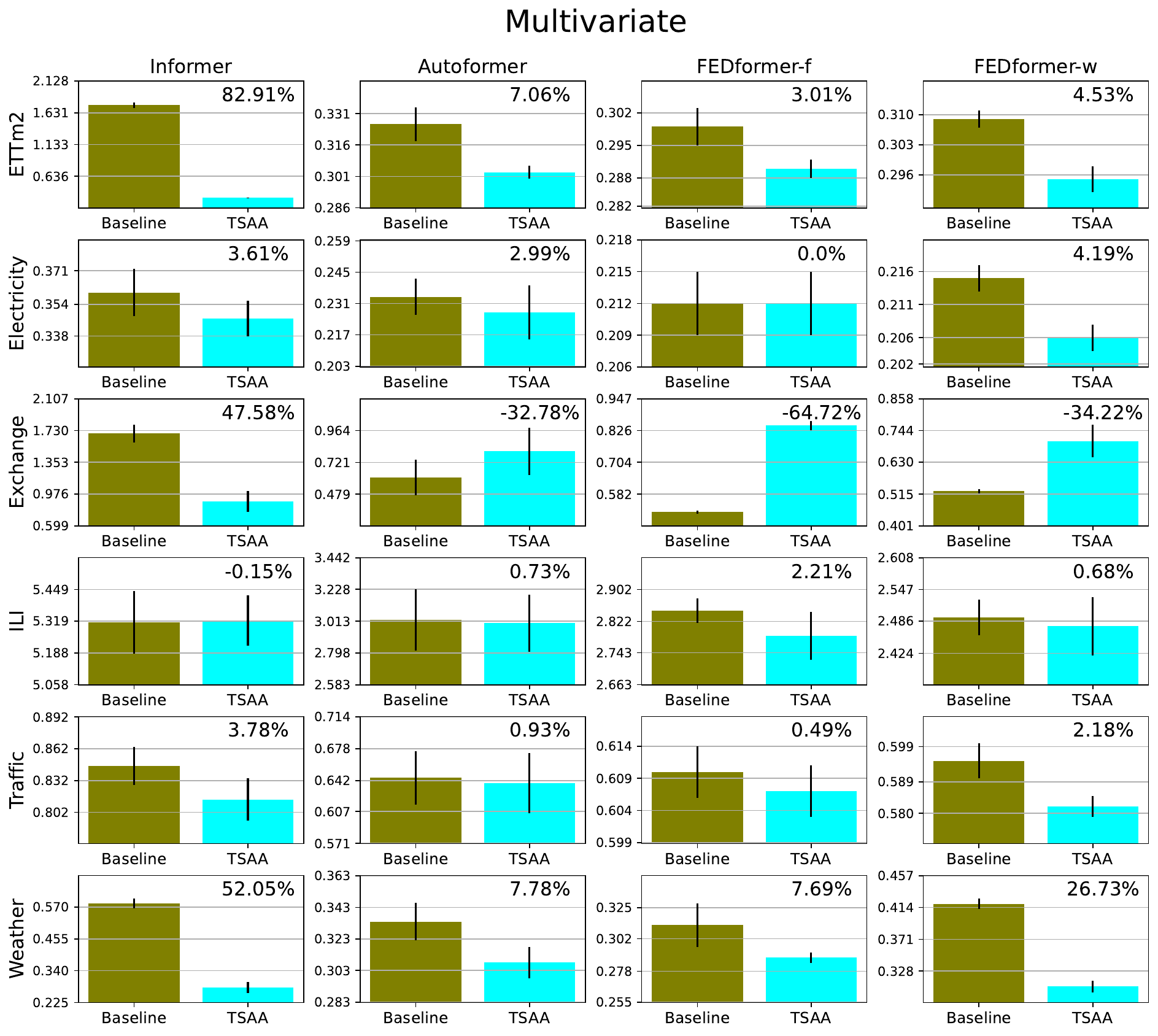}
    \caption{Multivariate comparison between TSAA and the baseline models per dataset. Rows represent the datasets and columns represent the models. Each score represents the average MSE across all four horizons, namely, 96,192,336, and 720.}
    \label{fig:final_table_plots}
\end{figure}

\begin{figure}[t]
    \centering
    \includegraphics[width=1.0\textwidth]{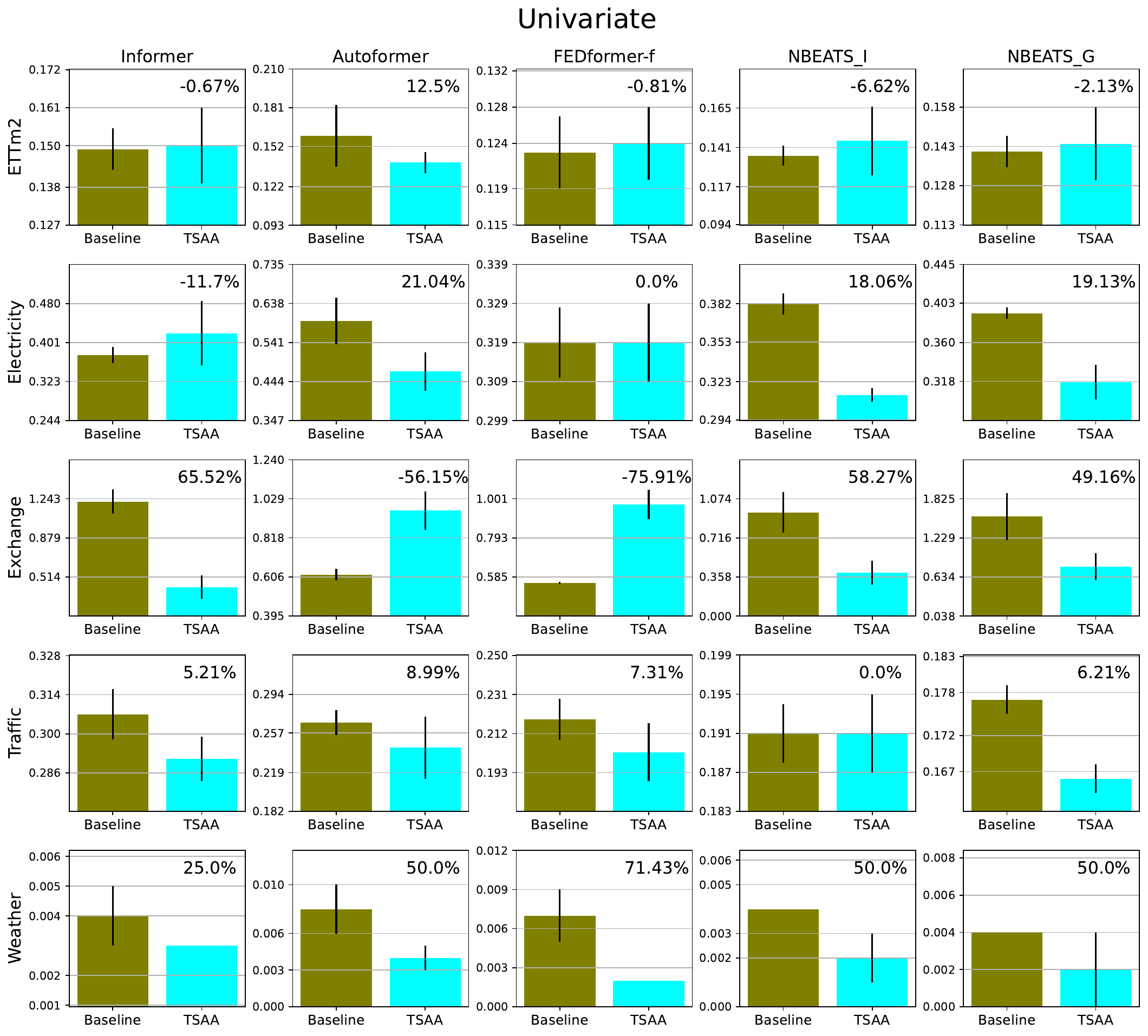}
    \caption{Univariate comparison between TSAA and the baseline models per dataset. Rows represent the datasets and columns represent the models. Each score represents the average MSE across all four horizons, namely, 96,192,336, and 720.}
    \label{fig:final_table_plots_uni}
\end{figure}

\clearpage

\begin{table*}[!h]
 \centering
    \caption{Informer: Comparison of automatic augmentation approaches including TSAA, Fast AutoAugment and RandAugment.}
    \label{tab:aa_results_informer}
    \scalebox{0.6}{
\begin{tabular}{ll|cc|cc|cc|cc}
\toprule
Dataset &  Horizon &      Baseline MSE &      Baseline MAE &   RandAugment MSE &   RandAugment MAE &       Fast AA MSE &       Fast AA MAE &        TSAA MSE &          TSAA MAE \\
\midrule
  \multirow{4}{*}{\rotatebox[origin=c]{90}{ETTm2}} &   96 &             0.545 &             0.588 & \underline{0.417} &             0.498 &             0.443 & \underline{0.488} &  \textbf{0.224} &    \textbf{0.321} \\
   &      192 &             1.054 &             0.808 & \underline{0.583} & \underline{0.579} &             0.796 &             0.661 &   \textbf{0.27} &    \textbf{0.355} \\
   &      336 &             1.523 &             0.948 & \underline{0.957} & \underline{0.757} &             1.565 &             0.974 &  \textbf{0.304} &    \textbf{0.374} \\
   &      720 &             3.878 &             1.474 & \underline{1.266} & \underline{0.886} &             3.639 &             1.469 &  \textbf{0.398} &    \textbf{0.435} \\ \hline
    \multirow{4}{*}{\rotatebox[origin=c]{90}{ECL}} &   96 &             0.336 & \underline{0.416} & \underline{0.325} &             0.418 &             0.369 &             0.445 &  \textbf{0.324} &    \textbf{0.407} \\
     &      192 &              0.36 &             0.441 & \underline{0.347} & \underline{0.435} &             0.388 &             0.463 &  \textbf{0.336} &    \textbf{0.419} \\
     &      336 &             0.356 & \underline{0.439} & \underline{0.354} &             0.441 &             0.916 &             0.776 &  \textbf{0.347} &    \textbf{0.429} \\
     &      720 &             0.386 & \underline{0.452} & \underline{0.385} &             0.453 &             1.016 &             0.818 &  \textbf{0.381} &    \textbf{0.448} \\ \hline
\multirow{4}{*}{\rotatebox[origin=c]{90}{Traffic}} &   96 & \underline{0.744} &  \underline{0.42} &             0.754 &             0.425 &             0.948 &             0.539 &  \textbf{0.723} &    \textbf{0.408} \\
 &      192 & \underline{0.753} & \underline{0.426} &              0.78 &             0.434 &             1.446 &             0.777 &  \textbf{0.735} &    \textbf{0.414} \\
 &      336 & \underline{0.876} & \underline{0.495} &             0.926 &              0.52 &             1.484 &             0.814 &  \textbf{0.811} &    \textbf{0.462} \\
 &      720 & \underline{1.011} & \underline{0.578} &              1.12 &             0.632 &             1.511 &             0.819 &  \textbf{0.985} &    \textbf{0.566} \\ \hline
\multirow{4}{*}{\rotatebox[origin=c]{90}{Weather}} &   96 &             0.315 &             0.382 &             0.489 &             0.485 & \underline{0.258} & \underline{0.339} &   \textbf{0.18} &    \textbf{0.256} \\
 &      192 &             0.428 & \underline{0.449} &             0.468 &             0.488 & \underline{0.415} &             0.456 &  \textbf{0.253} &    \textbf{0.331} \\
 &      336 &              0.62 &             0.554 &             0.585 &             0.551 & \underline{0.542} & \underline{0.519} &  \textbf{0.296} &    \textbf{0.361} \\
 &      720 &             0.975 &             0.722 &               1.0 &             0.741 & \underline{0.819} & \underline{0.665} &  \textbf{0.392} &    \textbf{0.426} \\ \hline
    \multirow{4}{*}{\rotatebox[origin=c]{90}{ILI}} &   24 &             5.349 &             1.582 &     \textbf{4.82} &    \textbf{1.451} & \underline{5.046} & \underline{1.505} &           5.313 &             1.559 \\
     &       36 & \underline{5.203} &             1.572 &    \textbf{4.326} &    \textbf{1.385} &             5.264 & \underline{1.549} &            5.26 &             1.581 \\
     &       48 & \underline{5.286} & \underline{1.594} &    \textbf{4.655} &    \textbf{1.456} &             5.449 &             1.596 &           5.415 &             1.623 \\
     &       60 &             5.419 &              1.62 &    \textbf{4.542} &    \textbf{1.448} &             5.684 &             1.659 & \underline{5.3} & \underline{1.593} \\
\bottomrule
\end{tabular}}
\end{table*} 

\begin{table*}[!h]
 \centering
    \caption{Autoformer: Comparison of automatic augmentation approaches including TSAA, Fast AutoAugment and RandAugment.}
    \label{tab:aa_results_autoformer}
    \scalebox{0.6}{
\begin{tabular}{ll|cc|cc|cc|cc}
\toprule
Dataset &  Horizon &      Baseline MSE &      Baseline MAE &   RandAugment MSE &   RandAugment MAE &       Fast AA MSE &       Fast AA MAE &          TSAA MSE &          TSAA MAE \\
\midrule
  \multirow{4}{*}{\rotatebox[origin=c]{90}{ETTm2}} &   96 &             0.231 &              0.31 &             0.223 &             0.307 & \underline{0.222} & \underline{0.303} &    \textbf{0.211} &    \textbf{0.293} \\
   &      192 &             0.289 &             0.346 &             0.286 &             0.343 & \underline{0.282} & \underline{0.334} &    \textbf{0.269} &    \textbf{0.327} \\
   &      336 &             0.341 &             0.375 & \underline{0.333} & \underline{0.371} &             0.352 &             0.377 &    \textbf{0.322} &    \textbf{0.359} \\
   &      720 &             0.444 &             0.434 &             0.427 &              0.42 & \underline{0.422} & \underline{0.413} &      \textbf{0.41} &    \textbf{0.407} \\ \hline
    \multirow{4}{*}{\rotatebox[origin=c]{90}{ECL}} &   96 &               0.2 &             0.316 & \underline{0.192} & \underline{0.307} &             0.213 &             0.321 &    \textbf{0.188} &    \textbf{0.302} \\
     &      192 &    \textbf{0.217} &    \textbf{0.326} &    \textbf{0.217} &    \textbf{0.326} &             0.249 &             0.346 & \underline{0.221} & \underline{0.328} \\
     &      336 &             0.258 &             0.356 &    \textbf{0.232} &    \textbf{0.341} &  \underline{0.25} &             0.354 &             0.252 & \underline{0.352} \\
     &      720 & \underline{0.261} & \underline{0.363} &             0.279 &             0.378 &             0.274 &              0.37 &    \textbf{0.248} &    \textbf{0.351} \\ \hline
\multirow{4}{*}{\rotatebox[origin=c]{90}{Traffic}} &   96 & \underline{0.615} & \underline{0.384} &             0.637 &             0.394 &             0.663 &             0.418 &    \textbf{0.602} &    \textbf{0.375} \\
 &      192 &              0.67 &             0.421 &    \textbf{0.636} &    \textbf{0.393} &             0.764 &              0.48 & \underline{0.663} & \underline{0.416} \\
 &      336 &             0.635 & \underline{0.392} & \underline{0.632} & \underline{0.392} &             0.711 &             0.443 &    \textbf{0.627} &    \textbf{0.387} \\
 &      720 &    \textbf{0.658} &    \textbf{0.402} &             0.663 &             0.409 &             0.872 &             0.526 & \underline{0.662} & \underline{0.405} \\ \hline
\multirow{4}{*}{\rotatebox[origin=c]{90}{Weather}} &   96 &             0.259 &             0.332 &             0.251 &             0.325 &    \textbf{0.196} &     \textbf{0.26} & \underline{0.216} & \underline{0.292} \\
 &      192 &             0.298 &             0.356 &             0.286 &             0.344 &    \textbf{0.255} &    \textbf{0.304} & \underline{0.278} & \underline{0.336} \\
 &      336 &             0.357 &             0.394 & \underline{0.328} & \underline{0.364} &    \textbf{0.307} &    \textbf{0.338} &             0.341 &             0.381 \\
 &      720 &             0.422 &             0.431 &             0.436 &             0.445 &    \textbf{0.375} &    \textbf{0.381} & \underline{0.397} &  \underline{0.41} \\ \hline
    \multirow{4}{*}{\rotatebox[origin=c]{90}{ILI}} &   24 &    \textbf{3.549} & \underline{1.305} &             3.774 &             1.354 &             4.737 &             1.603 & \underline{3.565} &    \textbf{1.302} \\
     &       36 & \underline{2.834} & \underline{1.094} &              2.89 &             1.103 &             3.864 &             1.375 &    \textbf{2.754} &    \textbf{1.068} \\
     &       48 &             2.889 &             1.122 &    \textbf{2.575} &    \textbf{1.045} &             3.766 &             1.368 & \underline{2.856} & \underline{1.114} \\
     &       60 & \underline{2.818} & \underline{1.118} &    \textbf{2.775} &    \textbf{1.108} &              3.92 &             1.412 &             2.826 & \underline{1.118} \\
\bottomrule
\end{tabular}}
\end{table*}

\begin{table*}[!h]
 \centering
    \caption{FEDformer-f: Comparison of automatic augmentation approaches including TSAA, Fast AutoAugment and RandAugment.}
    \label{tab:aa_results_fedformer}
    \scalebox{0.6}{
\begin{tabular}{ll|cc|cc|cc|cc}
\toprule
Dataset &  Horizon &      Baseline MSE &      Baseline MAE &   RandAugment MSE &   RandAugment MAE &       Fast AA MSE &       Fast AA MAE &          TSAA MSE &          TSAA MAE \\
\midrule
  \multirow{4}{*}{\rotatebox[origin=c]{90}{ETTm2}} &   96 & \underline{0.189} &             0.282 &             0.192 &             0.282 &             0.197 & \underline{0.279} &    \textbf{0.187} &    \textbf{0.274} \\
   &      192 &             0.258 &             0.326 & \underline{0.257} &             0.323 &             0.262 & \underline{0.319} &    \textbf{0.255} &    \textbf{0.314} \\
   &      336 &             0.323 &             0.363 &             0.326 &             0.364 & \underline{0.322} & \underline{0.356} &    \textbf{0.311} &     \textbf{0.35} \\
   &      720 &             0.425 &             0.421 &              0.43 &             0.422 & \underline{0.415} & \underline{0.407} &    \textbf{0.406} &    \textbf{0.403} \\ \hline
    \multirow{4}{*}{\rotatebox[origin=c]{90}{ECL}} &   96 &    \textbf{0.185} &      \textbf{0.3} &  \underline{0.19} & \underline{0.304} &             0.201 &             0.313 &    \textbf{0.185} &      \textbf{0.3} \\
     &      192 &    \textbf{0.201} & \underline{0.316} &             0.206 &             0.318 & \underline{0.203} &    \textbf{0.315} &    \textbf{0.201} & \underline{0.316} \\
     &      336 &    \textbf{0.214} &    \textbf{0.329} &             0.227 &             0.338 &  \underline{0.22} & \underline{0.331} &    \textbf{0.214} &    \textbf{0.329} \\
     &      720 &    \textbf{0.246} &    \textbf{0.353} &             0.291 &             0.385 & \underline{0.254} & \underline{0.357} &    \textbf{0.246} &    \textbf{0.353} \\ \hline
\multirow{4}{*}{\rotatebox[origin=c]{90}{Traffic}} &   96 &    \textbf{0.577} &    \textbf{0.361} &  \underline{0.59} &             0.376 &             0.655 &              0.41 &    \textbf{0.577} & \underline{0.362} \\
 &      192 &  \underline{0.61} & \underline{0.379} &             0.622 &              0.39 &             0.652 &             0.408 &    \textbf{0.601} &    \textbf{0.371} \\
 &      336 & \underline{0.623} & \underline{0.385} &             0.626 &             0.392 &             0.674 &             0.421 &    \textbf{0.619} &    \textbf{0.383} \\
 &      720 &    \textbf{0.632} &    \textbf{0.388} & \underline{0.643} & \underline{0.396} &             0.705 &             0.427 &    \textbf{0.632} &    \textbf{0.388} \\ \hline
\multirow{4}{*}{\rotatebox[origin=c]{90}{Weather}} &   96 &             0.236 &             0.316 & \underline{0.203} & \underline{0.275} &    \textbf{0.191} &    \textbf{0.252} &             0.207 &             0.285 \\
 &      192 &             0.273 &             0.333 &             0.267 &             0.331 &     \textbf{0.24} &     \textbf{0.29} & \underline{0.252} & \underline{0.311} \\
 &      336 &             0.332 &             0.371 &             0.329 &             0.368 &     \textbf{0.29} &    \textbf{0.321} & \underline{0.313} & \underline{0.355} \\
 &      720 &             0.408 &             0.418 &             0.398 &             0.413 &    \textbf{0.363} &    \textbf{0.367} & \underline{0.382} & \underline{0.395} \\ \hline
    \multirow{4}{*}{\rotatebox[origin=c]{90}{ILI}} &   24 &             3.268 &             1.257 &  \underline{3.16} & \underline{1.234} &             4.671 &             1.612 &     \textbf{3.15} &    \textbf{1.219} \\
     &       36 &             2.648 &             1.068 &    \textbf{2.457} &    \textbf{1.019} &             3.835 &             1.377 & \underline{2.578} & \underline{1.049} \\
     &       48 &             2.615 &             1.072 &    \textbf{2.558} &     \textbf{1.06} &             3.694 &             1.359 & \underline{2.609} & \underline{1.069} \\
     &       60 &             2.866 &             1.158 & \underline{2.822} & \underline{1.146} &             3.855 &              1.41 &    \textbf{2.805} &     \textbf{1.14} \\
\bottomrule
\end{tabular}}
\end{table*}

\end{document}